\documentclass[final,onefignum,onetabnum]{siamart190516}

\usepackage{lipsum}
\usepackage{amsfonts}
\usepackage{graphicx}
\usepackage{epstopdf}
\usepackage{algorithmic}
\ifpdf
  \DeclareGraphicsExtensions{.eps,.pdf,.png,.jpg}
\else
  \DeclareGraphicsExtensions{.eps}
\fi

\newsiamremark{remark}{Remark}
\newsiamremark{hypothesis}{Hypothesis}
\crefname{hypothesis}{Hypothesis}{Hypotheses}
\newsiamthm{claim}{Claim}

\headers{Hawkes Processes Modeling, Inference and Control: An Overview}{R. Lima}

\title{Hawkes Processes Modeling, Inference and Control: An Overview\thanks{Preprint. Work in Progress.
\funding{This work was not supported by any organization.}}}

\author{Rafael Lima\thanks{Samsung R\&D Institute Brazil, Campinas-SP, Brazil 
  (\email{rafael.goncalves.lima at gmail.com}).}}

\usepackage{amsopn}

\ifpdf
\hypersetup{
  pdftitle={Hawkes Processes Modeling, Inference and Control: An Overview},
  pdfauthor={R. Lima}
}
\fi

\usepackage{booktabs} 
\usepackage{bbold}
\usepackage{mathrsfs}
\usepackage{subcaption}
\usepackage{amssymb}
\usepackage{amsmath}
\usepackage{hyperref}
\usepackage{mathtools}
\usepackage{xcolor,colortbl}
\usepackage{color}

\DeclareMathOperator*{\argmax}{argmax}
\DeclareMathOperator*{\argmin}{argmin}

\DeclarePairedDelimiter\floor{\lfloor}{\rfloor}

\usepackage{graphicx}

\begin{document}

\maketitle

\begin{abstract}
Hawkes Processes are a type of point process which model self-excitement among time events. It has been used in a myriad of applications, ranging from finance and earthquakes to crime rates and social network activity analysis. Recently, a surge of different tools and algorithms have showed their way up to top-tier Machine Learning conferences. This work aims to give a broad view of the recent advances on the Hawkes Processes modeling and inference to a newcomer to the field. The parametric, nonparametric, Deep Learning and Reinforcement Learning approaches are broadly discussed, along with the current research challenges on the topic and the real-world limitations of each approach. Illustrative application examples in the modeling of Retweeting behaviour, Earthquake aftershock occurence and COVID-19 spreading are also briefly discussed.
\end{abstract}



\section{Introduction}

Point Processes are tools for modeling the arrival of time events. They have been broadly used to model both natural and social phenomena related to arrival of events in a continuous time-setting, such as the queueing of customers in a given store, the arrival of earthquake aftershocks\cite{YO99,AH02}, the failure of machines at a factory, the request of packages over a communication network, and the death of citizens in Ancient societies \cite{VJ03}.

Predicting, and thus being able to effectively intervene in all these phenomena is of huge commercial and/or societal value, and thus there has been an intensive investigation of the theoretical foundations of this area.

Hawkes Processes (HP) \cite{AH71} are a type of point process which models self- and mutual-excitation, i.e., when the arrival of an event makes future events more likely to happen. They are suitable for capturing epidemic, clustering, and faddish behaviour on social and natural time-varying phenomena. The excitation effect is represented by an additional function to the intensity of the process (i.e., the expected arrival rate of events): the triggering kernel, which quantifies the influence of events of a given process in the self- and mutual triggering of its associated intensity functions. Much of the Hawkes Processes' research has been devoted to modeling the triggering kernels, handling issues of scalability to large number of concurrent processes and quantity of data, as well as speed and tractability of the inference procedure.

Regarding the learning of the triggering kernels, one of the methods involves the assumption that it can be defined by simple parametric functions, s.a. one or multiple exponentials, Gaussians, Rayleigh, Mittag-Leffler \cite{JC20} functions and power-laws. Much of the work dealing with this type of approach concerned with enriching these parametric models \cite{RK16,YW16,HX16,HX18,ND15}, scaling them for high dimensions, i.e., multivariate processes of large dimensions \cite{EB15,RL17}, dealing with distortions related to restrictions on the type of available data \cite{HX17}, and proposing adversarial losses as a complement to the simple Maximum Likelihood Estimation (MLE) \cite{JY18}.

Another way of learning the triggering function is by assuming that it is represented by a finite grid, in which the triggering remains constant along each of its subintervals. Regarding this piecewise constant (or non-parametric) approach, most notably developed in \cite{GM12,EB16}, the focus has been on speeding up the inference through parallelization and online learning of the model parameters \cite{YY17,MA17}.

A more recent approach, enabled by the rise to prominence of Deep Learning models and techniques, which accompany the increase of computational power and availability of data of the recent years, regards modeling the causal triggering effect through the use of neural network models, most notably RNNS, LSTMs and GANs. These models allow for less bias and more flexibility than the parametric models for the modeling of the triggering kernel, while taking advantage of the numerous training and modeling techniques developed by the booming connectionist community.

In addition, regarding the control of self-exciting point processes,i.e., the modification of the process parameters towards more desirable configurations, while taking into account an associated `control cost' to the magnitude of these modifications, recent works either make use of Dynamic Programming (Continuous Hamilton-Jacobi-Bellman Equation-based approach) \cite{AZ17,AZ172}, Kullback-Leibler Divergence penalization (a.k.a. `Information Bottleneck') \cite{YW17}, and Reinforcement Learning-based, as well as Imitation Learning-based techniques \cite{UU18,SL18}.

Although there have been some interesting reviews and tutorials regarding domain-specific applications of Hawkes Processes in Finance \cite{AH18,EB15} and Social Networks \cite{MR18}, a broad view of the inference and modeling approachs is still lacking. A close work to ours is the one presented in \cite{JY19} which, although very insightful, lacks coverage of important advances, such as the previously mentioned control approaches, as well as the richer variants of Neural Network-based models. Furthermore, a concise coverage of the broader class of Temporal Point Processes is given in \cite{MG18}, while reviews for parametric spatiotemporal formulations of HP are given in \cite{AR18} and \cite{BY18}.

In the following, we introduce the mathematical definitions involving HPs, then carefully describe the advances on each of the aforementioned approachs, then finish by a summarization, along with some considerations.

\section{Theoretical Background}
In this and the next section, we present the mathematical definitions used throughout the remaining sections of the paper. The Hawkes Processes were originally introduced in \cite{AH71} and \cite{AH712}.

In the present work, we restrain ourselves to the Marked Temporal Point Processes (MTPP), i.e., the point process in which each event is defined by a time coordinate and a mark (or label). An intuitive example of a MTPP is shown in Figure \ref{fig:mtpp_counting_process}.

The key definitions are those of: Counting Process, Intensity Function, Triggering Kernel, Impact Matrix, (Log-)likelihood, Covariance, Bartlet Spectrum, Higher-order Moments and Branching Structure.

\subsection{Multivariate Marked Temporal Point Processes}

Realizations of univariate MTPPs, here referred to by $\mathcal{S}$ are one or more sequences of events $e_i$, each composed by the time coordinate $t_k$ and the mark $m_k$, s.a.:
\begin{equation}
\mathcal{S} = \{ (t_0,m_0), ... ,(t_S,m_S) \},
\end{equation}
where $S$ is the total number of events. Marks may represent, for example, a specific user in a social network or a specific geographic location, among others. For more complex problems, as in the check-in times prediction of \cite{GY18}, a composite mark may represent an user of interest and a specific location.

\begin{figure}[ht]
\begin{subfigure}{0.49\textwidth}
\centering
\includegraphics[width=\linewidth]{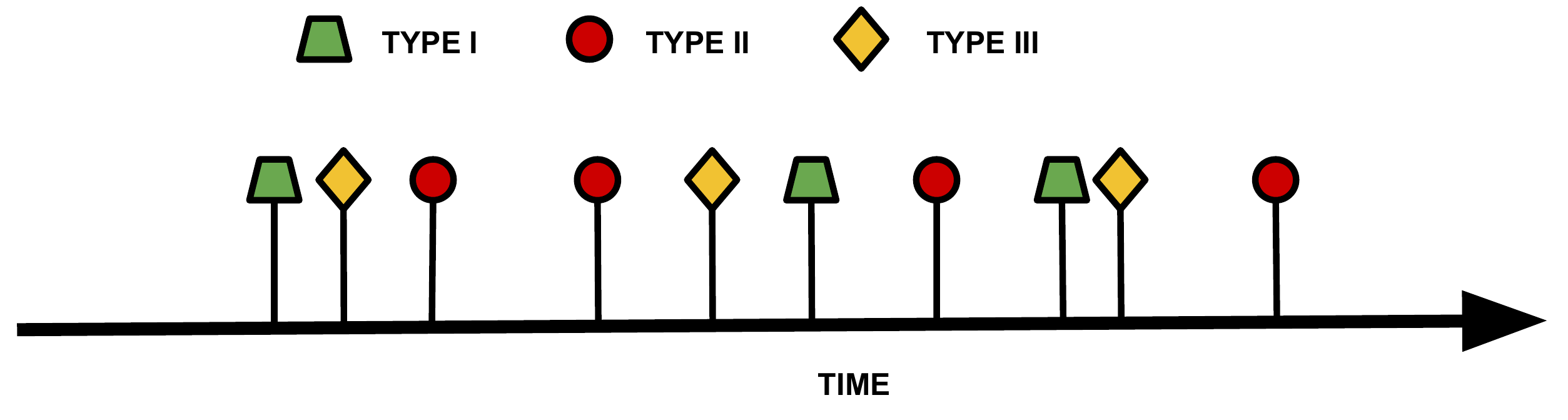} 
\label{fig:pp_example}
\end{subfigure}
\begin{subfigure}{0.49\textwidth}
\centering
\includegraphics[width=\linewidth]{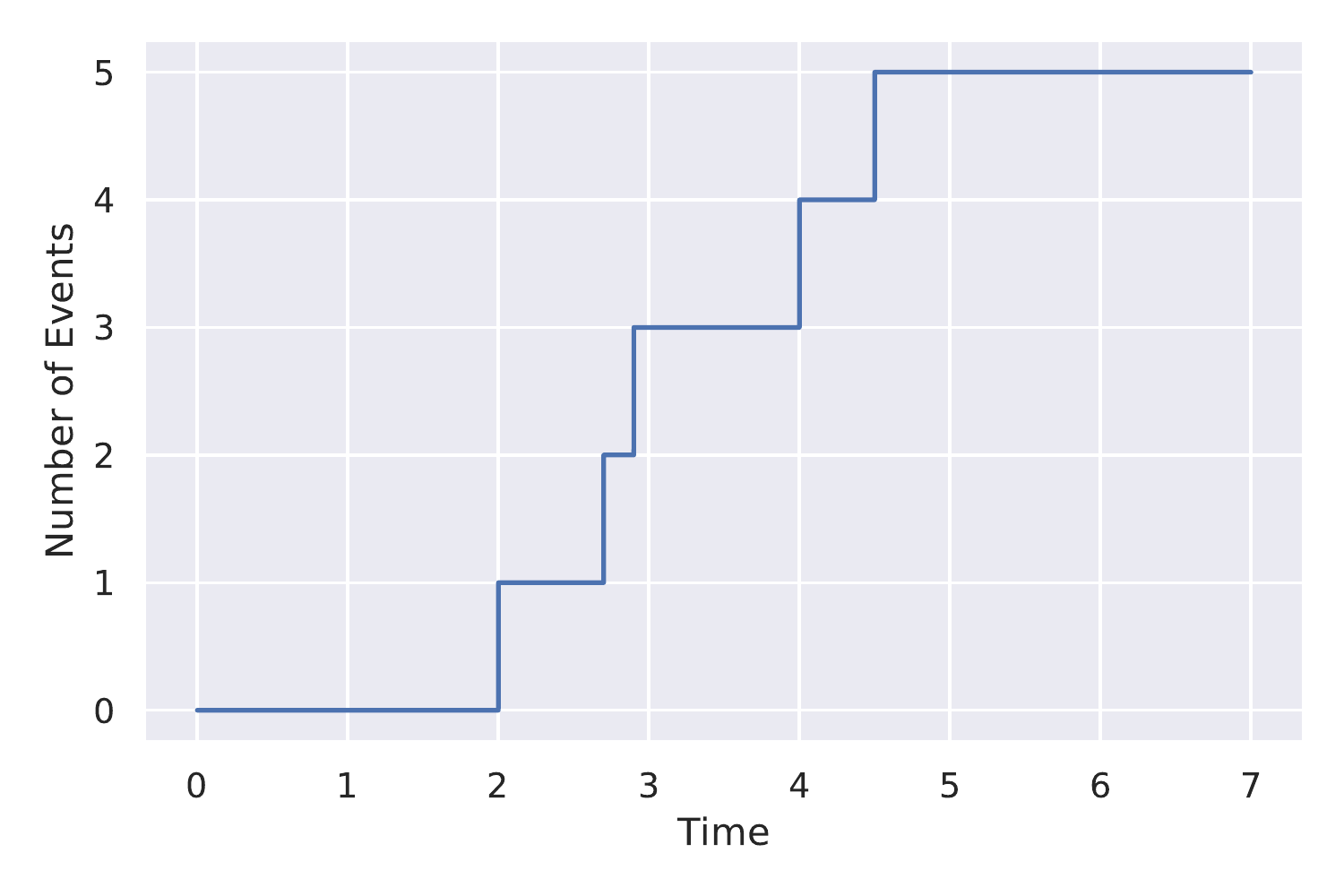}  
\label{fig:counting_process}
\end{subfigure}
\caption{Left: Intuitive diagram of a Marked Temporal Point Process (MTPP) with three types of event (marks). Right: Example of a counting process on the time interval [0,7].}
\label{fig:mtpp_counting_process}
\end{figure}

An easy way to generalize this notation would be to refer to multiple realizations of multivariate MTPPs as $\mathbf{\mathcal{S}} = \{\mathcal{S}_{i,j}\}$, where $i$ would refer to the dimension of the process, while $j$ would refer to the index of the sequence.
Now, regarding only the purely temporal portion of the process, i.e., the time coordinates $t_k$, it is also common to express them by means of a Counting Process $N(t)$, which is simply the cumulative number of event arrivals up to time t:
\begin{equation}
\int_{0^{-}}^{t} dN_s,
\end{equation}
where:
\begin{itemize}
\item $dN_{t_k} = 1$, if there is an event at $t_k$;
\item $dN_{t_k} = 0$, otherwise.
\end{itemize}
This is illustrated in Figure \ref{fig:mtpp_counting_process}.

Associated to each temporal point process, there is an Intensity Function, which is the expected rate of arrival of events:
\begin{equation}
\lambda (t) dt= E\left\{dN_t = 1\right\},
\end{equation}
which may depend or not on the history of past events. Such dependence results in a so-called ``Conditional Intensity Function'' (CIF):
\begin{equation}
\lambda (t)dt = E\left\{dN_t = 1 | \mathcal{H}\right\},
\end{equation}
where $\mathcal{H}$ is the history of all events up to time t:
\begin{equation}
\mathcal{H}: \left\{t_{i,j} \in \mathbf{\mathcal{S}} | t_{i,j} < t\right\}
\end{equation}
This concept will be further discussed in the next section.

\subsection{Hawkes Processes}

The simplest example of a temporal point process is the Homogeneous Poisson Process (HPP), in which the intensity is a positive constant:
\begin{equation}
\lambda (t) = \mu, 
\end{equation}
for $\mu \in \mathbb{R}^{+}$.

\begin{figure}[ht]
\begin{subfigure}{0.49\textwidth}
  \centering
  \includegraphics[width=\linewidth]{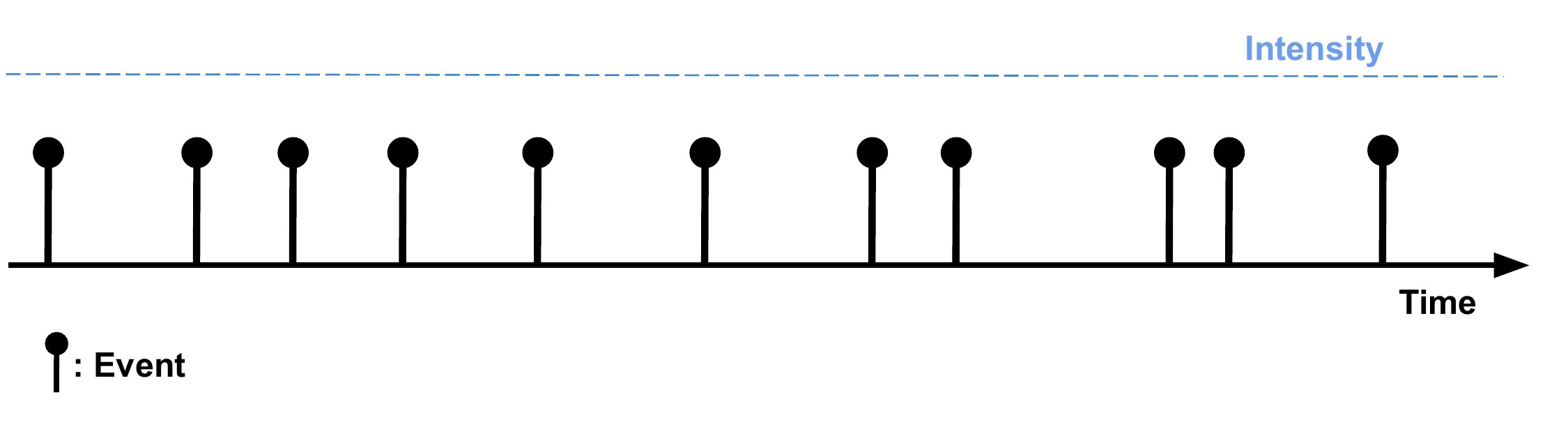}  
  \label{fig:sub-first}
\end{subfigure}
\begin{subfigure}{0.49\textwidth}
  \centering
  \includegraphics[width=\linewidth]{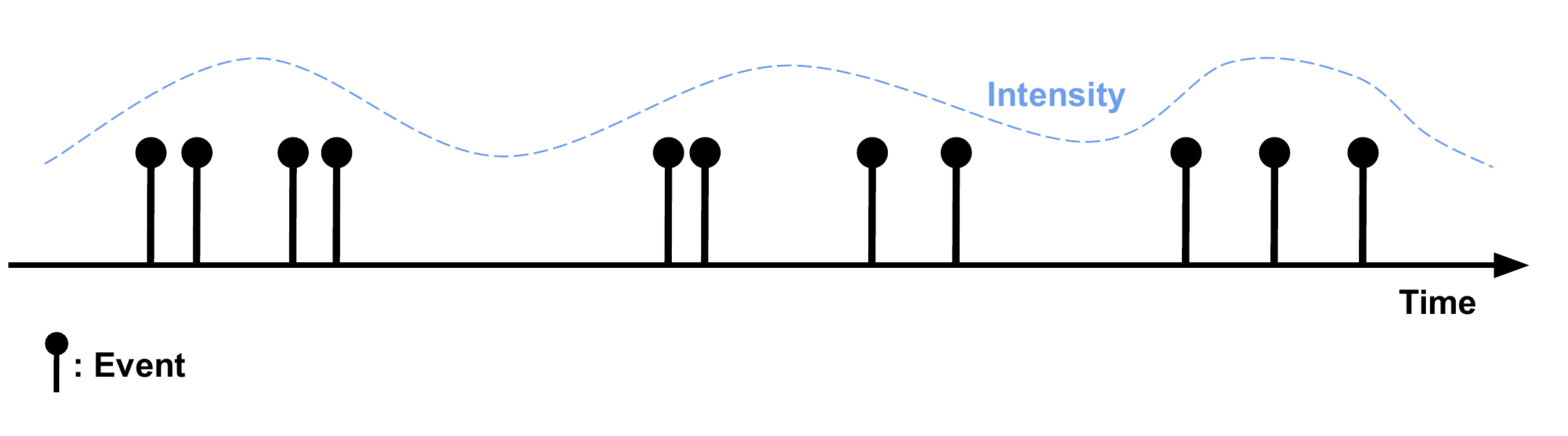}  
  \label{fig:sub-second}
\end{subfigure}
\caption{Illustrative examples of Homogeneous (left) and Inhomogeneous (right) Poisson Processes. The events are represented by black dots.}
\label{fig:ipp_hpp}
\end{figure}

In the case of an Inhomogeneous Poisson Process (IPP), the intensity $\lambda (t)$ is allowed to vary. Both HPP and IPP are shown in Figure \ref{fig:ipp_hpp}.

Both the Homogeneous and the Inhomogeneous Poisson Processes have, in common, the fact that each consecutive event interval sampled from the Intensity Function is independent of the previous ones. When analyzing several natural phenomena, one may wish to model how events in each dimension $i$ of the process,  which may be representing a specific social network user, an earthquake shock at a given geographical region, or a percentual jump in the price of a given stock, just to cite a few examples - affect the arrival of events in all the dimensions of the process, including its own.

In particular, we are interested in the cases where the arrival of one event makes further events more likely to happen, which is reflected as an increase in the value of the intensity function after the time of said event. When this increase happens to the Intensity Function of the same $i$-th dimension of the event, the effect is denominated self-excitation. When the increase happens to the intensity of other dimensions, we refer to it as mutual excitation.

Hawkes Processes (HP) model self-excitation in an analytical expression for the intensity through the insertion of an extra term, which is designed to capture the effect of all the previous events of the process in the current value of the CIF. For a univariate Hawkes, we have:
\begin{equation}
\lambda_{HP} (t)= \underbrace{\mu}_\textrm{baseline intensity} + \underbrace{\sum_{t_i < t} \phi(t-t_i)}_\textrm{self-excitation term},
\label{eq: intensity_HP}
\end{equation}
while, for the multivariate case, with dimension D, we are going to have both self-excitation ($\phi_{ii} (t)$) and mutual-excitation terms ($\phi_{ij} (t)$ , s.t.  $(i \neq j)$):
\begin{equation}
\lambda_{HP}^{i} (t) = \mu_i + \sum_{j=1}^{D} \sum_{t_{ij} < t} \phi_{ij} (t-t_{ij}).
\end{equation}
The assumptions of \begin{enumerate}
    \item Causality: $$\phi (t) = 0 \hspace{2em} \forall t < 0$$
    \item Positivity: $$\phi (t) \geq 0 \hspace{2em} \forall t \geq 0$$
\end{enumerate}
are usually held for all $\phi_{ij} (t)$ \footnote{The case in which $\phi (t) < 0 \text{ for } t \geq 0$ is referred to as an ``Inhibiting Process'', and not usually considered in Hawkes Processes works.}. 

In the case that the kernel matrix $\boldsymbol{\Phi} (t) = \left[\phi_{ij} (t)\right]_{i,j = 0}^{d,n}$
can be factored into $\boldsymbol{\Phi} (t) = \boldsymbol{\alpha} \odot \boldsymbol{\kappa} (t)$, with $\boldsymbol{\alpha} = \left[\alpha_{ij}\right]_{i,j = 0}^{d,n}$ and $\boldsymbol{\kappa} (t) = \left[\kappa_{ij} (t)\right]_{i,j = 0}^{d,n}$, where ``$\odot$'' correspond to the Hadamard (element-wise) product.

\begin{figure}[ht]
\begin{subfigure}{0.49\textwidth}
  \centering
  \includegraphics[width=\linewidth]{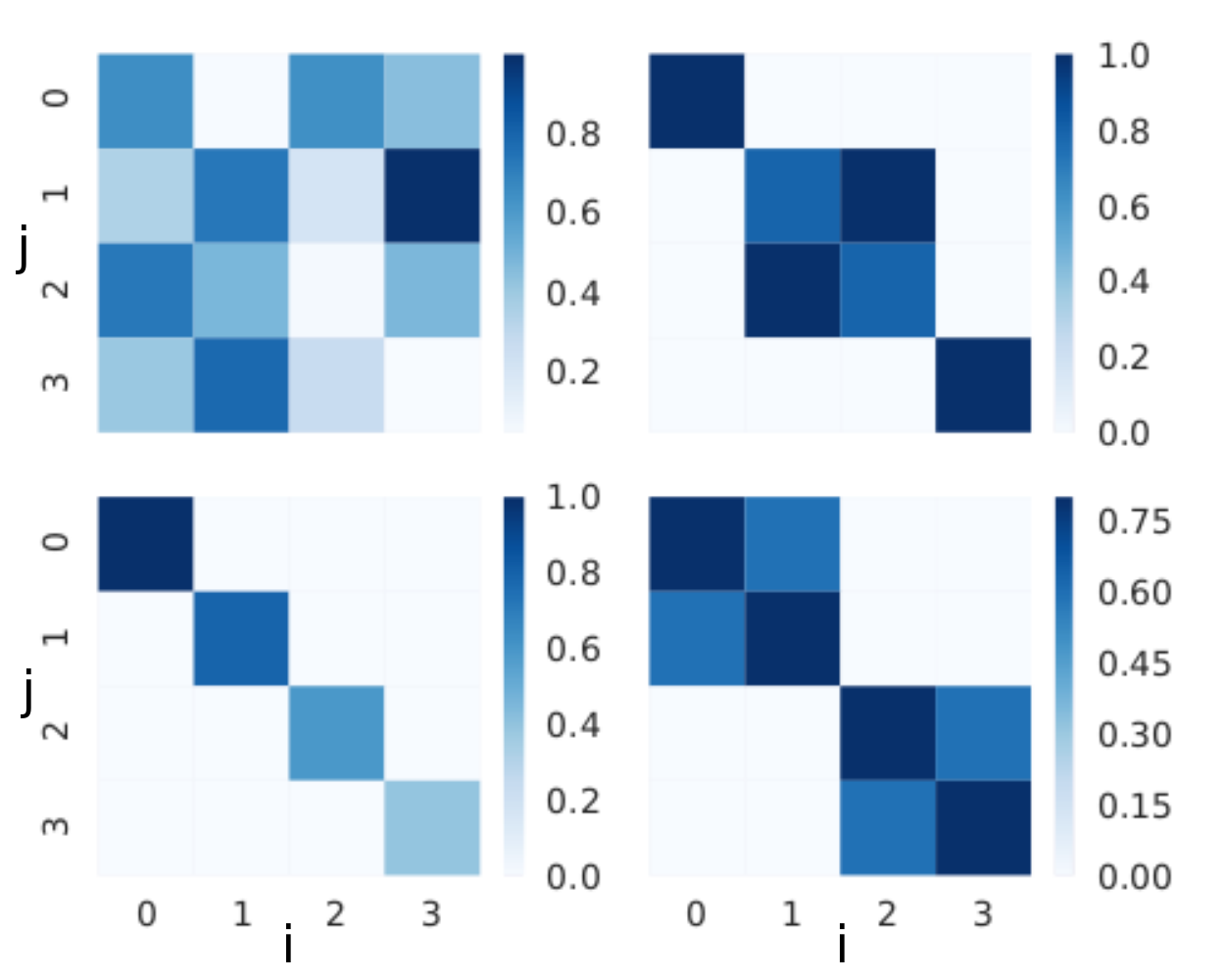}  
  \label{fig:impact_matrices}
\end{subfigure}
\begin{subfigure}{0.49\textwidth}
  \centering
  \includegraphics[width=\linewidth]{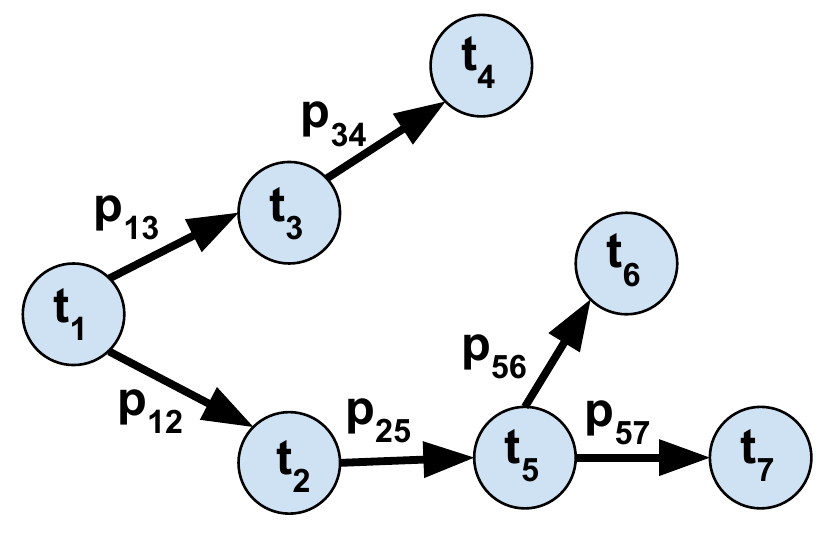}  
  \label{fig:branching_structure}
\end{subfigure}
\caption{Left: Four examples of $4 \times 4$ Impact Matrices $\boldsymbol{\alpha}$. Each $\alpha_{ij} (t)$ has the corresponding value indicated according to the color scale on the side. Right: Illustrative example of the concept of Branching Structure. A given edge $t_i \rightarrow t_j$ means that $t_i$ triggered $t_j$ with the probability $p_ji$.}
\label{fig:impact_matrices_branching_structure}
\end{figure}

 $\boldsymbol{\alpha}$, here denominated Impact Matrix, can implicitly capture a myriad of different patterns of self- and mutual excitation, as exemplified in Figure \ref{fig:impact_matrices_branching_structure}. This factorization is particularly convenient when adding penalization terms related to network properties to the loglikelihood-based loss of a Multivariate HP, such as in the works of \cite{HX16} and \cite{YL182}.


For being of practical value, realizations of HPs are constrained to having finite number of events for any sub-interval of the simulation horizon $[0,T]$. This corresponds to having the kernel function (or kernel matrix) satisfying the following stationarity condition:
\begin{equation}
\boldsymbol{Spr} \left(\left|\left| \boldsymbol{\Phi} (t) \right|\right| \right)\\ =
 \boldsymbol{Spr} \left(\{\left|\left|\phi_{ij} (t) \right| \right|\}_{1 \leq i,j\leq D} \right) \\ < 1,
\end{equation}
where $||\phi(t)||$ corresponds to $||\phi(t)|| = \int_0^{\infty} \phi (t) dt$ and $\boldsymbol{Spr} (\cdot)$ corresponds to the spectral radius of the matrix, i.e., the largest value among its eigenvalues.

If this stationarity condition is satisfied, we have that the process will reach weakly stationary state, i.e., when the properties of the process, most notably its ``moments'', vary only as a function of the relative distance, here referred to as ``$\tau$'', of its points.

The first order moment, or statistics of the HP, is defined as:
\begin{equation}
\Lambda_i = E\{\lambda_{HP}^{i} (t)\} = \lim_{T \to \infty} \dfrac{1}{T} \int_{0}^{T} \lambda_{HP}^{i} (t) dt = {(\mathbb{I} - ||\boldsymbol{\Phi (t)}||)}^{-1}\mu_i
\end{equation}
while the second-order statistics, or stationary covariance, is defined as:
\begin{equation}
\nu^{ij} (t'-t) dt dt' = E \{dN_t^i dN_{t'}^j \} - \Lambda_i \Lambda_j dt dt' - \epsilon_{ij} \Lambda_i \delta(t'-t) dt,
\end{equation}
where $\epsilon_{ij}$ is 1, if $i = j$, and 0, otherwise, while $\delta(t)$ refers to the Dirac delta distribution.

The Fourier Transform of this stationary covariance is referred to as \textit{Bartlett Spectrum}. Sometimes, a different transform, the Laplace Transform, is used for the same purpose. In Hawkes' seminal paper \cite{AH71}, high importance is given to the fact that, when assuming some specific parametric functions for the excitation matrix $\boldsymbol{\Phi (t)}$, it is possible to find simple formulas for the covariance of process in the frequency domain. One example is the univariate case for $\phi (t)$ defined as the frequently used ``parametric exponential kernel'': $\phi (t) = \alpha e^{-\beta t}$, for $\alpha, \beta \in \mathbb{R}$.

For this choice, we have:
\begin{equation}
\nu^{*} (s) = \mathcal {L} \{ \nu\} (s)= \dfrac{\alpha \mu (2 \beta -\alpha)}{2 (\beta - \alpha) (s + \beta - \alpha)} \hspace{2em} (s \in \mathbb{C})
\end{equation}
where $\mathcal{L} \{ \cdot\} (s)$ refers to the Laplace Transform \footnote{The Laplace Transform $\mathcal{L} \{f\} (s)$ of a function $f(t)$ defined for $t \geq 0$ is computed as $\mathcal{L} \{f\} (s)= \int_{0^-}^{\infty} f(t) \mathrm{e}^{-s t}$, for some $s \in \mathbb{C}$.}. The detailed steps of this computation can be found in \cite{AH71}.

Going beyond the first- and second-order statistics, it is also possible to define statistics of higher orders, see \cite{TL09,VJ03}. Although they become less and less intuitive and tractable, as their order increases, the work in \cite{MA17} makes use of 3rd-order statistics, $K^{ijk}$, in a specific application of the Generalized Method of Moments for learning the impact matrix of multivariate HPs. It is defined as:
\begin{align}
K^{ijk} dt &= \int \int_{\tau, \tau' \in \mathbb{R}^2} \left(\mathbb{E} (dN_{t}^i dN_{t+\tau}^{j} dN_{t+\tau'}^{k}) \right. \\\nonumber &\left.- 2 \mathbb{E} (dN_{t}^i) \mathbb{E} (dN_{t+\tau}^{j}) \mathbb{E} (dN_{t+\tau'}^{k}) - \mathbb{E} (dN_t^i dN_{t+\tau}^j) \mathbb{E} (dN_{t+\tau'}^k) \right. \\\nonumber &\left.- \mathbb{E} (dN_t^i dN_{t+\tau'}^k) \mathbb{E} (dN_{t+\tau}^j) - \mathbb{E} (dN_{t+\tau}^j dN_{t+\tau'}^k) \mathbb{E} (dN_{t}^i)\right),
\end{align}
for $1 \leq i,j,k \leq D$, and it is connected to the skewness of $N_t$.

Now, regardless of the function family chosen for modeling $\boldsymbol{\Phi (t)}$ and $\boldsymbol{\mu}$, its fitness will be computed by measuring its likelihood over a set of sequences similar to the set of sequences used for training the model. Let be a set $\boldsymbol{\mathcal{S}}$ of M sequences, each with a total number of $N_j$ events, considered over the interval $\left[0,T\right]$, such that:
\begin{equation}
    \boldsymbol{\mathcal{S}} = \{ \mathcal{S}^j\}_{j=1}^M = \{\left[(t_1^j,m_0^j),...,(t_{N_j}^j,m_{N_j}^j)  \right]\}_{j=1}^M,
\end{equation}
with $m_j^k \in \{1,2,...,D\}$, $\forall j,k \in \mathbb{Z}_+$.

And let be a family $\mathbb{F}$ of Multivariate HPs with dimension $D \geq 1$ and parametric exponential kernels assumed for the shape of the excitation functions, such that the CIF $\lambda_{i}^{j} (t)$ of each i-th node defined over the sequence $\mathcal{S}^j$ is given by \footnote{Equivalent definitions of $\mathbb{F}$ can be given to families of HPs defined by other types of HPs, such as those with power-law kernels, or those with the corresponding CIF modeled by a recurrent neural network.}:
\begin{equation}
    \lambda_i^{j} (t) = \mu_i + \sum_{t_k^j \leq t} \alpha_{m_{k}^j i}\mathrm{e}^{-\beta_{m_{k}^j i}(t-t_{k}^j)} \hspace{2em} (t_k^j \in \mathcal{S}^j)
\end{equation}
Given parameter vectors $$\boldsymbol{\mu} = \{\mu_m\}_{m=1}^D \in \mathbb{R}_+^D \hspace{2em} \theta = \{(\alpha_{mn},\beta_{mn})\}_{m=1,n=1}^{D,D} \in \mathbb{R}_+^{2D^2},$$ the likelihood function given in the logarithmic form, i.e., the loglikelihood, of a multivariate HP over a set $\mathcal{S}$ of M sequences considered over the interval $\left[0,T\right]$, is given by:
\begin{equation}
llh_{\mathcal{S}}(\boldsymbol{\mu}, \theta, \mathbb{F}) = \sum_{j=1}^M \left( \sum_{i=1}^{D} \sum_{k \in N_{j}} \log \lambda_i^{j} (t_k^j) - \underbrace{\sum_{i=1}^{D} \int_{0}^{T} \lambda_{i}^{j} (t) dt}_\textrm{Compensator} \right)
\end{equation}
 Thus, the goal of learning a HP over $\mathcal{S}$, is the act of finding vectors $\boldsymbol{\mu}$ and $\theta$ s.t.:
\begin{equation}
\label{eq: MLE}
(\boldsymbol{\mu},\theta) = \argmax llh_{\mathcal{S}}(\boldsymbol{\mu}, \theta, \mathbb{F})   
\end{equation}
A more rigorous and complete derivation of Equation \ref{eq: MLE} can be found in \cite{TB11}. 

Another concept which is of relevance in some inference methods is that of a Branching Structure ($\mathcal{B}$). It defines the ancestry of each event in a given sequence, i.e., specifies the probability that the i-th event $t_i$ was caused by the effect of a preceding event $t_j$ in the CIF ($p_{ji}\text{, for } 0 \leq i < n$) or by the baseline intensity $\mu$ ($p_{0i}$). The probabilities $p_{ji}$ and $p_0i$ can be given by:
\begin{equation}
p_{ji} = \dfrac{\phi(t_i-t_j)}{\lambda(t_i)}, \text { for } (j \geq 1) \hspace{1em} \text{and} \hspace{1em} p_{0i} = \dfrac{\mu}{\lambda(t_i)}.
\end{equation}
As an example, consider the example illustrated in Figure \ref{fig:impact_matrices_branching_structure}, in which event $t_1$, the first of the event series, causes events $t_2$ and $t_3$; event $t_2$ causes event $t_5$; event $t_3$ causes event $t_4$; and event $t_5$ causes events $t_6$ and $t_7$. The corresponding Branching Structure $\mathcal{B}_E$ implied by these relations among the events has an associated probability given by:
\begin{equation}
    p(\mathcal{B}_E) = p_{01}*p_{12}*p_{13}*p_{25}*p_{34}*p_{56}*p_{67}
\end{equation}


\subsection{Simulation Algorithms}

Regarding the experimental aspect of HPs, synthetic data may be generated through the following methods:
\begin{enumerate}
    \item \textbf{Ogata's Modified Thinning Algorithm} \cite{YO81}: It starts by sampling the first event at time $t_0$ from the baseline intensity. Then, each posterior event $t_i$ is obtained by sampling it from a HPP with intensity fixed as the value calculated at $t_{i-1}$, and then:
    \begin{itemize}
        \item Accepting it with probability $\dfrac{\lambda(t_i)}{\lambda^*(t_{i-1})}$, where $\lambda^* (t_{i-1})$ is the value of the intensity at time $t_{i-1}$, while $\lambda_{t_i}$ is the value calculated through Equation \ref{eq: intensity_HP}.
        \item Or rejecting it and proceed to resampling a posterior event candidate;
    \end{itemize}
    \item \textbf{Perfect Simulation} \cite{JR10}: It derives from the fact the HP may be seen as a superposition of Poisson Processes. It proceeds by sampling events from the baseline intensity, taken as the initial level, and then sampling levels of descendant events for each of the events sampled at the previous level. From each event $t_{0,i}$ sampled from the baseline intensity, we associate an IPP with intensity defined as $\phi(t-t_{0,i})$, and then sample its descendant events. Next, we take each descendant event sampled and also associate it with its corresponding IPP, and so on, until all the levels were explored over the simulation horizon $[0,T]$.
\end{enumerate}

\begin{algorithm}
\caption{Ogata's Modified Thinning Algorithm (Univariate Case)}
\label{alg:ogata_algo}
\begin{algorithmic}
\STATE{Input $\mu$, $\phi (t)$, T}
\STATE{Define t = 0}
\STATE{Sample $t_1$ from exponential distribution with rate $\mu$}
\STATE{Update $t = t + t_1$}
\STATE{Define $n = 1$}
\WHILE{$t < T$}
\STATE{$\lambda_{n} = \mu + \sum_{i=1}^{n} \phi(t - t_{n})$}
\STATE{Sample $t_{n+1}$ from exponential distribution with rate $\lambda_{n}$}
\STATE{$\lambda_{n+1} = \mu + \sum_{i=1}^{n} \phi(t + t_{n+1} - t_{n})$}
\STATE{Sample $u$ from Uniform Distribution over $\left[0,1\right]$}
\IF{$\dfrac{\lambda_{n+1}}{\lambda_{n}} < u$}
\STATE{Update $ t = t+t_{n+1}$}
\STATE{Update $n = n+1$}
\ENDIF
\ENDWHILE
\RETURN{$\{t_i\}_{i=1}^{n}$}
\end{algorithmic}
\end{algorithm}

\begin{algorithm}
\caption{Perfect Simulation of Hawkes Processes (Univariate Case)}
\label{alg:perfect_simulation}
\begin{algorithmic}
\STATE{Input $\mu$, $\phi (t)$, T}
\STATE{Define j = 0}
\STATE{Simulate HPP with $\lambda=\mu$ over $\left[0,T\right]$ to obtain $\{t_{j}^{i}\}_{i=1}^{n_j}$}
\WHILE{$\exists t_{j}^{i} \hspace{1em} (\forall i \leq n_j)$}
\FOR{($i = 1$ ; $i \leq n_j$ ; $i++$)}
\STATE{Simulate IPP with $\lambda=\phi(t-t_{j}^{i})$ over $t \in \left[0,T\right]$ to obtain $\{t_{(j+1),k}^i\}_{k=1}^{n_{j+1}^i}$}
\ENDFOR
\STATE{Update $n_{j+1} = \sum_{i=1}^{n_j} n_{j+1}^i$}
\STATE{Update $\{t_{(j+1)}^{i}\}_{i=1}^{n_{j+1}} = \bigcup_{i=1}^{n_{j+1}} \{t_{(j+1),k}^i\}_{k=1}^{n_{j+1}^i}$}
\STATE{Update $j = j+1$}
\ENDWHILE
\RETURN{$\bigcup_{l=0}^{j} \{t_{l}^i\}_{i=1}^{n_{l}}$ \hspace{1em} (After sorting)}
\end{algorithmic}
\end{algorithm}

More detailed, step-by-step descriptions of each of these two algorithms are shown in pseudocode format, in Algorithm \ref{alg:ogata_algo} and Algorithm \ref{alg:perfect_simulation}, both for the case of Univariate HPs. In the next section, we focus on the HP models which assume simple parametric forms for the excitation functions, along with its variants, which we will refer to as ``Parametric HPs''.

\section{Parametric HPs}
\label{sec: param}

\begin{figure*}
    \centering
    \includegraphics[width=\textwidth]{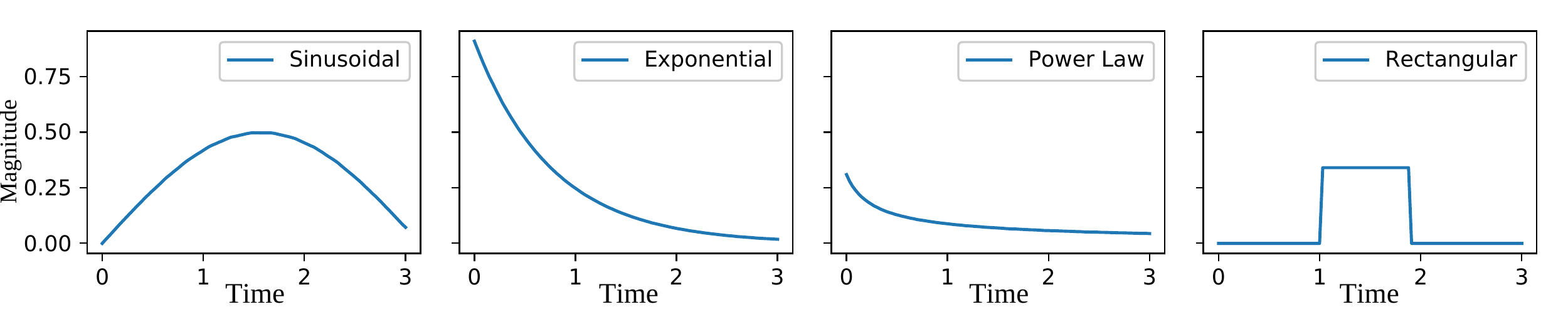}
    \caption{Four examples of parametric HP kernels ($\phi (t)$). Each of them is used to model a different type of interaction among events of a given HP.}
    \label{fig:hp_kernels}
\end{figure*}

In the present section, we discuss the HP models which assume simple parametric forms for the excitation functions. Figure \ref{fig:hp_kernels} shows some examples of commonly used functions for parametric HPs. Much has been done recently in terms of incrementing these models for dealing with specific aspects of some domains, such as social networks \cite{QZ15,RK16}, audio streaming \cite{YW16}, medical check-ups \cite{HX17}, among others. In the following subsections, we aim to give a broad view of how the parametric modeling of HPs are used and improved throughout a series of possible ideas. We divided the recent research on parametric HPs as focusing on three different strategies, accompanied by working examples. The referred strategies are:
\begin{enumerate}
    \item Enhancing and composing simple parametric kernels, to adapt the model to specific modeling situations and datasets;
    \item Improving scalability of parametric HP models, to the multivariate cases with many nodes and sequences with many jumps;
    \item Improving robustness of training over worst-case scenarios and defective data.
\end{enumerate}

Further examples on each strategy are also briefly mentioned in Section \ref{sec: current_challenges}.

\subsection{Enhanced and Composite Triggering Kernels}

As a way of modeling the daily oscillations of the triggering effects on Twitter data, \cite{RK16} proposes a time-varying excitation function for HPs. The probability $\mathbb{P}$ of getting a retweet over the time interval $[t, t+\delta t]$, with small $\delta$, is modeled as:
\begin{equation}
\mathbb{P} (\text{Retweet in }[t, t + \delta t]) = \lambda(t) \delta t,
\end{equation}
in which the time-dependent rate is dependent on previous events as:
\begin{equation}
\lambda (t) = p(t) \sum_{t_i < t} d_i \phi (t - t_i),
\end{equation}
with $p(t)$ being the infectiousness rate, $t_i$ as the time corresponding to the i-th retweet arrival, and $d_i$ as the number of followers of the i-th retweeting individual.

Furthermore, the memory kernel $\phi (s)$, a probability distribution for the time intervals
between a tweet by the followee and its retweet by the
follower, has been shown to be heavily
tailed in a variety of social networks \cite{QZ15}. It is
fitted to the empirical data by the function:
\[
\phi(s)=
\left\{
\begin{array}{lr}
0,& \text{ for } s < 0 \\
c_0 ,& \text{ for } 0 \leq s \leq s_0 \\
c_0 (s/s_0)^{-(1+\theta)},& \text{ for } s > s_0
\end{array}
\right.
\]
where the parameters $c_0$, $s_0$ and $\theta$ are known.

The model is defined so that the daily cycles of human activity are naturally translated into cycles of retweet activity. The time dependence of the infectious rate is, therefore, defined as:
\begin{equation}
p(t) = p_0 \left\{ 1 - r_0 \sin{\left(\dfrac{2 \pi}{T_m} (t + \phi_0)\right)}\right\}\sqrt[\tau_m]{e^{-(t-t_0)}}
\end{equation}

The parameters $p_0$, $r_0$, $\phi_0$ and $\tau_m$ correspond to the intensity, the relative amplitude of the oscillation, its phase, and the characteristic time of popularity decay respectively. Those are fitted through a Least Square Error (LSE) minimization procedure over the aggregation of retweet events over time bins $\delta t$.

Another improvement over the traditional parametric froms for HPs involve the addition of a nonlinearity on the expression for the CIF:
\begin{equation}
\lambda_{HP} (t) = g \left( \mu + \sum_{t_i < t} \phi(t-t_i)\right),
\end{equation}
in which $g(\cdot)$ correspond to a so-called \textit{link function}, e.g., sigmoid:
\begin{equation}
g(x) = \dfrac{1}{1+e^{-x}}.
\end{equation}
The work in \cite{YW16} proposes a procedure for simultaneously learning $g(\cdot)$, $\mu$ and $\phi (t) = \alpha \kappa(t)$, with $\kappa(t)$ taken as $e^{-t}$, for assuring convergence, of the algorithm.

The procedure is formulated using a moment-matching idea over a piecewise-constant approximation for $g(\cdot)$, which leads to the definition of the objective function as a summation:
\begin{equation}
\min_{g \in \mathcal{G}, \boldsymbol{W}} \dfrac{1}{n} \sum_{i=1}^{n} \left(N_i - \int_{0}^{t_i} g(w \cdot x_t ) dt\right)^2,
\end{equation}
with $w = (\mu,\alpha)^T$, and $x_t = (1, \sum_{t_i \in \mathcal{H}_t} \kappa (t - t_i))^T$.

The algorithm is run by recursively updating the estimates $\hat{w}$, with the Isotron Algorithm (see \cite{AK09}), and $\hat{g}$, with a projected gradient descent step.

Theoretical bounds for the approximation error of the method are also given, along with extensions of the algorithm for general point processes, monotonically decreasing nonlinearities, low-rank processes and multidimensional HPs.

Another possible enhancement for modeling the parametric HP is through using a composition of Gaussian kernels of different bandwidths, as in \cite{HX16}. The core idea is that the maximum nonzero frequency component of the kernel is bounded as the same value of the intensity function, since this one is simply a weighted sum of the basis functions. Therefore, for every value of the tolerance $\xi$, it is possible to find a frequency value $\omega_0$ s.t.:
\begin{equation}
\int_{\omega_0}^{\infty} |\hat{\lambda} (\omega)| d \omega \leqslant \xi 
\end{equation}
From this $\omega_0$, the method then defines the triggering function $\phi (t)$ as a composition of $\tilde{D}_{\phi}$ Gaussian functions, with $\tilde{D}_{\phi}$ equally spaced values of bandwidth over the interval $[0,\omega_0]$.

The estimate of the value of the intensity function, for transforming into the frequency domain, is done through a kernel density estimation with gaussian kernels of bandwidth fixed as the Silverman's rule of thumb:
\begin{equation}
h = \sqrt[5]{\left( \dfrac{4 \hat{\sigma}^5}{3 n}\right)} \approx \dfrac{1.06 \hat{\sigma}}{\sqrt[5]{n}},
\end{equation}
where $\hat{\sigma}$ is the standard deviation of the time intervals, and $n$ is the number of events of a given sequence.

After the Gaussian functions are defined, it remains to estimate the model coefficients $\Theta$, with the impact matrix, now an impact tensor $A = \{\alpha_{ijk}\}$, through a convex surrogate loss penalized by parameters related to the sparsity of the matrix, the temporal sparsity of the kernels, and the pairwise similarity:
\begin{equation}
\label{granger_caus_loss}
\argmin_{\Theta \geqslant 0} - \mathcal{L}_{\Theta} + \gamma_{\mathcal{S}} ||A||_1 + \gamma_{\mathcal{G}} ||A||_{1,2} + \gamma_{\mathcal{P}} E(A),
\end{equation}
where:
\begin{itemize}
\item  $||A||_1 = \sum_{i,j,k} |\alpha_{ijk}|$ is the l1-norm of the tensor, which is related to its temporal sparsity, which causes the excitation functions to go to zero at infinity, therefore maintaining the stability of the process;
\item $||A||_{1,2} = \sum_{i,j} ||\{\alpha_{ij1}, ..., \alpha_{ij\tilde{D}_{\phi}}\}||_2$ is related to the sparsity over the $\tilde{D}_{\phi}$ basis functions of a given node of the process, and it enforces the local independence of the process; 
\item $E(A) = \sum_{i}\sum_{i' \in \mathcal{C}_i} ||\boldsymbol{\alpha}_i - \boldsymbol{\alpha}_{i'}||_F^2 + ||\boldsymbol{\alpha}^i - \boldsymbol{\alpha}^{i'}||_F^2$ is a coefficient to enforce the pairwise similarity of the process, in which $\mathcal{C}_i$ corresponds to the cluster to which node i belongs, $||\cdot||_F$ is the Frobenius norm, $\boldsymbol{\alpha}_i = \{\alpha_{ijk}\}$, for fixed i, and $\boldsymbol{\alpha}^i = \{\alpha_{ijk}\}$, for fixed j. It means that, if $i$ and $i'$ are similar types of events, then their mutual excitation effects should be similar as well;
\item $\gamma_{\mathcal{S}}$, $\gamma_{\mathcal{G}}$ and $\gamma_{\mathcal{P}}$ are coefficients to be tuned for the model.
\end{itemize}
The estimation is done through an Expectation-Maximization procedure close to those of \cite{GM12} and \cite{KZ13}, which first randomly initializes the impact tensor and the vector of baseline intensities $\boldsymbol{\mu}$, and then iterates through:
\begin{enumerate}
\item Estimating the probability that each event was generating by each of the compositional basis kernels, as well as the baseline intensity;
\item Averaging the probabilities over all events of all training sequences for updating the coefficients of each basis function and the baseline intensity.
\end{enumerate}
The two steps are repeated until convergence of the parameter estimates.

\subsection{Scalability}

Another setting in which the parametric choice of kernels is highly convenient is with respect to the scalabity of the inference procedure for high dimensional networks and sequences with large number of events, which occur in several domains, such as social interactions data, which is simultaneously large (i.e., large number of posts), high-dimensional (numerous users) and structured (i.e., the users interactions are not in a random fashion but, instead, present some regularities).

One interesting inference method towards this direction is the work presented in \cite{RL17}, which achieves a complexity $O(n D)$, with D as the number of events comprised by the process history, and D as the dimension of the impact matrix of the process.

The referred method, entitled \textit{Scalable Low-Rank Hawkes Processes} (SLRHP), takes advantage of the memoryless property of the exponential and the underlying regularity of large networks connected to social events: The memoryless property, which means that, in HPs with exponential excitation functions, the effect of all past events over the intensity value of a given point can be computed by just the time of the last event before said point, speeds up the intensity computing portion of the inference procedure iterations, while the underlying regularity of large impact matrices associated with the social phenomena allow the dynamics of large-dimensional HPs to be captured by impact matrices of much smaller magnitudes.

The baseline rates and excitation functions of model are then defined using a low-rank approximation:
\begin{equation}
\mu_i = (t) = \sum_{j = 1}^{E} P_{ij} \tilde{\mu}_j
\end{equation}
\begin{equation}
\phi_{mi} (t) = \sum_{j,l = 1}^{E} P_{ij} P_{ml}\tilde{ \phi}_{lj} (t),
\end{equation}
in which $P \in \mathbb{R}_{+}^{D \times E}$ is a projection matrix from the original D-dimensional space to a low-dimensional space E ($E << D$). This projection can also be seen as a low-rank approximation of the excitation function matrix $\boldsymbol{\Phi}$, in which:
\begin{equation}
\boldsymbol{\Phi} = P \boldsymbol{\tilde{\Phi}} P^T
\end{equation}
Through having that $E << D$, the formulated low-rank approximated inference algorithm SLRHP manages to:
\begin{enumerate}
\item Capture a simplified underlying regularity impose inferred intensity rates' parameters by adopting sparsity-inducing constraints to the model parameters;
\item Lower the number of parameters for both the baseline rates and excitation kernels, with the D
natural rates and $D^2$ triggering kernels are lowered to r and
$E^2$, respectively. This advantage is diminished slightly by the additional cost of inferring the $(D \times E)$-sized projection matrix P.
\end{enumerate}

Another way of dealing with scalability issues of multivariate HPs, both in terms of the total number of events in the sequences and the number of nodes, is through the mean-field treatment, as described in \cite{EB153}.  Compared with the SLRHP method, which focuses on reducing the dimensionality of underlying network, the mean-field treatment focuses on finding closed-form expressions for approximate estimations involved in the optimization defined over the network in its real size. The key step of this method is to consider that the arrival intensities of each node of the process is wide-sense stationary, which implies the stability condition for the excitation matrix, and fluctuates only slightly around its mean value.

This last assumption, entitled \textit{Mean-Field Hypothesis}, posits that, if $\lambda^i (t)$ corresponds to the intensity of the i-th node, and $\tilde{\Lambda}^i$ corresponds to the empirical estimator of the first-order statistics of said node,
\begin{equation}
\tilde{\Lambda}^i = \dfrac{N_T^i}{T},
\end{equation}
where $N_T^i$ corresponds to the total number of events arrived at node i up to the final time of the simulation horizon $[0,T]$, then we have that:
\begin{equation}
\label{eq:mfh}
 \dfrac{|\lambda^i (t) - \tilde{\Lambda}^i|}{\tilde{\Lambda}^i} << 1 \hspace{4em} \forall t \in \left[0,T\right]
\end{equation}
The condition defined by Equation \ref{eq:mfh} is met when: (i) $||\phi(t)|| << 1$, independently of the shape of $\phi (t)$; (ii) when the dimensionality of the MHP is sufficiently high; and also (iii) when $\phi (t)$ changes sufficiently slowly, so that the influence of past events average to a near constant value. 

From this, we can recover the parameters $\theta^i$ from the intensity function $\lambda_t^i$, and the intensity function from the first-order statistics $\tilde{\Lambda^i}$, as:
\begin{equation}
\log \lambda_t^i \simeq \log \tilde{\Lambda}^i + \dfrac{\lambda^i (t)- \tilde{\Lambda}^i}{\tilde{\Lambda}^i} - \dfrac{(\lambda^i (t) - \tilde{\Lambda}^i)^2}{2 (\tilde{\Lambda}^i)^2}
\end{equation}
and
\begin{equation}
\lambda^i (t) = \mu^i + \int_{0^-}^{t} \sum_{j = 1}^D \phi^{ji} (t) dN_t^i
\end{equation}
The method yields mean-field estimates for the parameters with error which decays proportionally to the inverse of the final time T of the sequences:
\begin{equation}
\mathbb{E} (\theta^i) \approx \theta_{MF}^i
\end{equation}
\begin{equation}
\boldsymbol{cov} (\theta^{ji},\theta^{j'i'}) \sim \dfrac{1}{T},
\end{equation}
where $\theta_{MF}^i$ refers to the mean-field estimator of the parameters.

\subsection{Training}

Elaborating further on some difficulties of inferring parametric kernels from real-data, an interesting method regarding truncated sequences is described in \cite{HX17}. 

In the case of learning HP parameters from real-data, one often has to deal with sequences which are only partially observed, i.e., the time event arrivals are only available over a finite time-window. 

This poses a challenge concerning the robustness of the learning algorithms, since the triggering pattern from unobserved events are not considered: The inference deals with the error induced by computing intensity values over finite time windows, i.e., by computing intensity values along simulation horizons $[0,T]$, the closed-form equation would assume that its value at 0 is simply the baseline rate $\mu$. 

From the expression for the CIF:
\begin{equation}
\lambda_{HP} (t) = \mu + \sum_{t_i < t} \phi (t - t_i),
\end{equation}
for $t \in [0,T]$, we have that, taking some value $T' \in [0,T]$, we may split the triggering effect term in two parts:
\begin{equation}
\lambda_{HP} (t) = \mu + \sum_{t_i < T'} \phi (t - t_i) + \sum_{T' \leq t_i < T} \phi (t - t_i),
\end{equation}
for $t \in [T',T]$.

If we are observing the sequences only over the interval $[T',T]$, the second term is implicitly ignored, what may have severe degradation of the learning procedure, specially in the case that the excitation functions decay slowly.

The method proposed handles this issue through a \textit{sequence-stitching method}. The trick is to sample ``candidate predecessor events'' and choosing the most likely one from their similarity w.r.t. the observed events. The augmented sequences would then be use for the actual HP parameters' learning algorithm.

In practice, this means that, given M sequences $\mathcal{S}_m$ realized over $[T',T]$, the method would \textbf{not} learn from the regular MLE formula
\begin{equation}
\theta^* = \argmax_{\theta} llh (\{\mathcal{S}_m\}_{m=1}^{M},\theta),
\end{equation}
but, instead, from some expression which takes into account the expected influence of unobserved predecessor events
\begin{equation}
\theta_{SDC}^* = \argmax_{\theta} \mathbb{E}_{s ~ \mathcal{H}_{T'}} llh ([\mathcal{S},\mathcal{S}_m],\theta),
\label{eq: mle_sdc} 
\end{equation}
where $\mathcal{H}_{T'}$ corresponds to the distribution over all possible sequences of events happening before time $T'$. In practice, this expectation is computed not over the real distribution, but over some finite number of (relatively few) samples, such as 5 or 10. This finite sample approximation converts Equation \ref{eq: mle_sdc} onto
\begin{equation}
\theta_{SDC}^* = \argmax_{\theta} \sum_{\mathcal{S}_{stitch} \in \mathcal{K}} p (\mathcal{S}_{stitch}) llh (\mathcal{S}_{stitch},\theta),
\end{equation}
where $p (\mathcal{S}_{stitch})$ is the probability of the stitched sequence obtained from the concatenation of the original observed sequence and one of the sample predecessor candidate sequence. This probability is obtained by normalizing over similarity values over the candidate sequences obtained from some similarity function $\mathbb{S} (\cdot,\cdot)$ of the form:
\begin{equation}
\mathbb{S} (\mathcal{S}_k,\mathcal{S}) = \sqrt[\psi]{{\rm e}^{-||f (\mathcal{S}_k) - f(\mathcal{S}) ||^2 }},
\end{equation}
where $f (\cdot)$ is some feature of the event sequence, and $\psi \in \mathbb{R}^+$ is some scale parameter.
By fixing the excitation pattern matrix $\boldsymbol{\kappa}$ as composed of exponentials $\kappa (t) = e^{-\beta t}$ and imposing sparsity constraint $ ||\boldsymbol{\alpha}||_1 = \sum_{i,j} |\alpha_{ij}|$ over the impact matrix, the equivalent problem,
\begin{equation}
\theta^* = \argmax_{\mu \geqslant 0, \boldsymbol{\alpha} \succ 0} \sum_{\mathcal{S}_{stitch} \in \mathcal{K}} p (\mathcal{S}_{stitch}) llh (\mathcal{S}_{stitch},\theta) + \gamma ||\boldsymbol{\alpha}||_1
\end{equation}
is shown to be solved through Expectation-Maximization updated for both $\mu$ and $\boldsymbol{\alpha}$. The method is more suitable for slowly-decaying excitation patterns, in which the influence of the unobserved events would be more prominent. In the case of exponentials with large values for the decay factor $\beta$, the improvement margins mostly vanish.

Another interesting improvement regarding the training procedure of MLE for HP parametric functions involves the complementary use of adversarial and discriminative learning, as in \cite{JY18}. Although adversarial training has gained an ever increasing relevance for neural-network based models in the last years, due to the popularization of Generative Adversarial Networks \cite{IG14} and its variants, as will be discussed further in Section \ref{sec: nn-based}, keeping the assumption of a simple parametric shape for the excitation function is a way to insert domain-specific knowledge in the inference procedure.

The key idea of this complementary training is that, while the discriminative loss, here defined as the Mean-Squared Error (MSE) between discretized versions of predicted and real sequences, would direct the parameter updates towards smoother prediction curves, the adversarial loss would tend to push the temporally evenly distributed sampled sequences towards those more realistic-looking.

For the Gradient Descent-based updates, a discretization of the point process is carried out, so as to approximate the predictions through a recursive computation of the integral of the intensity function (the \textit{compensator} portion of the loglikelihood function) in a closed-form expression. The parameters of the complementary training are actually initialized by a purely MLE procedure, which was found to be more insensitive to initial points. The $MLE + GAN$ training updates then follows. The full procedure can be summarized by the following steps:
\begin{enumerate}
\item Subdivide the M original sequences on $[0,T]$ into training (on $[0,T^{tn}]$), validation (on $[T^{tn},T^{vd}]$) and test (on $[T^{vd},T]$) portions, using previously defined parameters $T^{tn}$ and $T^{vd}$;
\item Initialize the parameters of the model through the `purely' MLE procedure;
\item Sample M sequences from the model over the interval $[0,T^{tn}]$;
\item By choosing a specific parameter shape for the excitation function and binning both the original and simulated sequences over equally-spaced intervals, define the closed-form expression for the MSE (discriminative) loss $\mathcal{L}_{MSE}$ over all the sequences and dimensions of the process;
\item Define the GAN (adversarial) loss of the model over the sequences as:
\begin{equation}
\mathcal{L}_{GAN} =
\nonumber
\end{equation}
\begin{equation}
\begin{cases}
E_{\mathcal{S}^{tn} ~ \mathbb{P} (\mathcal{S}^{tn})} \left[F_W (Y_{\theta_{\mathcal{S}}} (\mathcal{S}^{tn})) \right] - E_{\mathcal{S}^{tn} ~ \mathbb{P} (\mathcal{S}^{tn})} \left[F_W (Y_{\theta_{\mathcal{S}}} (\mathcal{S}^{tn})) \right], \\ \text{ for the critic network $F_W$} \\
-E_{\mathcal{S}^{tn} ~ \mathbb{P} (\mathcal{S}^{tn})} \left[F_W (Y_{\theta_{\mathcal{S}}} (\mathcal{S}^{tn})) \right], \\ \text{ for the training model (generator)},
\end{cases}
\end{equation}
where $\mathbb{P} (\mathcal{S}^{tn})$ corresponds to the underlying probability distribution we assumed to have generated the original sequences $\mathcal{S}^{tn}$, $F_W \{\cdot\}$ refers to a neural network which can compute the so-called Wasserstein distance, a metric for difference among distributions which will be further explained in Section \ref{sec: nn-based}, and $Y_{\theta_{\mathcal{S}}} (\cdot)$ corresponds to the parametric model for sequence generation.
\item Compute the joint loss for MLE and GAN portions as:
\begin{equation}
\boldsymbol{L}_{MLE + GAN} = \gamma_{GAN} \boldsymbol{L}_{MLE} + (1-\gamma_{GAN}) \boldsymbol{L}_{GAN}, \text{ for }
 \gamma_{GAN} \in [0,1]
\end{equation}
\item Compute gradients of $\boldsymbol{L}_{MLE+GAN}$ over each parameter of model $Y_{\theta_{\mathcal{S}}}$;
\item Update parameter estimates of training model with $\eta \nabla_{\theta_{\mathcal{S}}} (\boldsymbol{L}_{MLE+GAN})$, for some learning rate $\eta$;
\item Repeat steps 3 to 8 until convergence.
\end{enumerate}
The method is an interesting combination of the enriched dynamical modeling from the adversarial training strategy with the robustness over small training sets of the parametric-based MLE estimation for HPs.

In this section, we provided a comprehensive analysis of the progresses in HP modeling and inference for excitation functions assumed to be of a simple parametric shape, along with their compositions and variants. In the next section, we will discuss about advances in nonparametric HP excitation function strategies.

\section{Nonparametric HPs}



Nonparametric HPs consider that some rigid and simple parametric assumption for the triggering kernel may not be enough to capture all the subtleties of the excitation effects that could not be retrieved from the data. They may be broadly divided between two main approaches:
\begin{enumerate}
    \item Frequentist
    \item Bayesian
\end{enumerate}
We will briefly discuss their variants in this section.

\subsection{Frequentist Nonparametric HPs}

The frequentist approach to HP modeling and inference consists in assuming that the excitation function (or matrix) can be defined as a binned grid (or a set of grids), in which the values of the functions would be taken as piecewise constant inside each bin, and the width of the bin will be (hopefully) expressive enough to model the local variations of the self-excitation effect.

They were first developed in the works of \cite{EL11}, \cite{EB12} and \cite{EB16}. In the case of \cite{EL11}, the final values of the bins were found by solving a discretized Ordinary Differential Equation, implied by the branching structure of the discretized triggering kernel and background rate over the data, through iterative methods. The approach in \cite{EB12} and \cite{EB16}, on the other side, recovers the piecewise constant model by exploiting relations, in the frequency domain, between the triggering kernel, the background rate and the second-order statistics of the model, also obtained in a discretized way over the data.

The increased expressiveness of this type of excitation model incurs in two main drawbacks:
\begin{itemize}
\item The bin division grid concept is close to that of a histogram over the distance among events, what usually requires much larger datasets for leading to accurate predictions, as opposed to parametric models, which would behave better on shorter and fewer sequences but most likely underfit on large sequence sets;
\item The time of the inference procedure may also be much larger, since it involves sequential binning computation procedures which can not take advantage of the markov property of parametric functions such as the exponentials. 
\end{itemize}

Two improvements, to be discussed in the present subsection, deal exactly with these drawbacks through:
\begin{itemize}
\item Acceleration of the computations over each sequence and/ or over each bin of the excitation matrix/function;
\item Reduction of the times of binning computational procedures through a so-called online update of the bin values.
\end{itemize}

\subsection{Acceleration of Impact Matrix Estimation through Matching of Cumulants}

One Acceleration strategy is developed in \cite{MA17}, which replaces the task of estimating the excitation functions directly through estimating their cumulative values, i.e., their integrated values from zero up to infinity, what would be enough to quantify the causal relationships among each node. That is, instead of estimating $\phi_{ij} (t)$, for each node, the method would estimate a matrix $\left|\left| \boldsymbol{\Phi} (t) \right|\right| = \{ \left|\left|  \phi_{ij} (t) \right|\right| \}$, in which:
\begin{equation}
\left|\left|  \phi_{ij} (t) \right|\right| = \int_{0}^{\infty} \phi_{ij} (t) dt\text{, } \forall (i,j) \in D \times D
\end{equation}
The method, entitled Nonparametric Hawkes Process Cumulant (NPHC), then proceeds to compute, from the sequences, moment estimates $\boldsymbol{\hat{R}}$ up to the third-order. It then finds some estimate $\left|\left| \hat{\boldsymbol{\Phi}} (t) \right|\right| $ of this cumulant matrix which minimizes the $L^2$ squared error between these estimated moments and the actual moments $\boldsymbol{R} (\left|\left| \boldsymbol{\Phi} (t) \right|\right|)$, which are uniquely determined from $\left|\left| \boldsymbol{\Phi} (t) \right|\right|$:
\begin{equation}
\left|\left| \hat{\boldsymbol{\Phi}} (t) \right|\right| = \argmin_{\left|\left| \boldsymbol{\Phi} (t) \right|\right|} ||\boldsymbol{R} (\left|\left| \boldsymbol{\Phi} (t) \right|\right|)-\boldsymbol{\hat{R}} ||^2
\end{equation}
This $L^2$ minimization comes from the fact that, by defining:
\begin{equation}
\boldsymbol{V} = (\mathbb{I}^D - \left|\left| \hat{\boldsymbol{\Phi}} (t) \right|\right|)^{-1},
\end{equation}
one may express the first-, second- and third-order moments of the process as:
\begin{equation}
\Lambda^i = \sum_{m=1}^D V^{im} \mu^m
\end{equation}
\begin{equation}
\nu^{ij} = \sum_{m=1}^D \Lambda^m V^{im} V^{jm}
\end{equation}
\begin{align}
K^{ijk} &= \sum_{m=1}^{D} \left(V^{im} V^{jm} \nu^{km} + V^{im} \nu^{jm} V^{km} \right. \nonumber\\
& \left. + \nu^{im} V^{jm} V^{km} - 2 \Lambda^m V^{im} V^{jm} V^{km} \right),
\end{align}
and thus we may find an estimator $\boldsymbol{\hat{V}} = \argmin_{\boldsymbol{V}} \boldsymbol{L}_{NPHC} (\boldsymbol{V})$, with $\boldsymbol{L}_{NPHC} (\boldsymbol{V})$ defined as:
\begin{equation}
\boldsymbol{L}_{NPHC} (\boldsymbol{V}) = (1-\gamma_{NPHC}) ||\boldsymbol{K^c} (\boldsymbol{V})-\boldsymbol{\hat{K^c}} ||_{2}^2 + \gamma_{NPHC}||\boldsymbol{\nu} (\boldsymbol{V}) - \boldsymbol{\hat{\nu}} ||_2^2,
\end{equation}
where $\gamma_{NPHC}$ is a weighting parameter, $|| \cdot ||_2^2$ is the Frobenius norm and $\boldsymbol{K^c} = \{K^{iij} \}_{1 \leq i,j \leq D}$ is a two-dimensional compression of the tensor $\boldsymbol{K}$. From this expression, by setting
\begin{equation}
\gamma_{NPHC} = \dfrac{||\boldsymbol{\hat{K^c}} ||_2^2}{||\boldsymbol{\hat{K^c}} ||_2^2 + || \boldsymbol{\hat{\nu}}||_2^2},
\end{equation}
one may arrive to
\begin{equation}
\left|\left| \hat{\boldsymbol{\Phi}} (t) \right|\right| = \mathbb{I}^D - \boldsymbol{\hat{V}}^{-1}
\end{equation}
The estimates of moments in the algorithm are actually computed through truncated and discretized (binned) countings along a single realization of the process and, since the real-data estimates are usually not symmetric, the estimates are averaged along positive and negative axis. 

Also, for $D = 1$, the estimate $\left|\left| \hat{\boldsymbol{\Phi}} (t) \right|\right|$ can be estimated solely from the second-order statistics. For higher-dimensional processes, it is the skewness of the third-order moment which uniquely fixes $\left|\left| \hat{\boldsymbol{\Phi}} (t) \right|\right|$.

\subsection{Online Learning}

Another improvement of the method consists in updating the parameters of the discretized estimate of the excitation function through a single pass over the event sequence, i.e., an online learning procedure \cite{YY17}.

In the case of the referred algorithm, the triggering function is assumed to:
\begin{enumerate}
\item be positive, 
\begin{equation}
\phi (t) \geq 0, \forall t \in \mathbb{R}
\end{equation}
\item have a decreasing tail, i.e.,
\begin{equation}
\sum_{k=m}^{\infty} (t_k - t_{k-1}) sup_{x \in (t_{k-1},t_k]} |f (y)| \leq \zeta_f (t_{i-1}), \forall i > 0,
\end{equation}
for some bounded and continuous $\zeta_f : \mathbb{R}^+ \mapsto \mathbb{R}^+$ s.t. $\lim_{t \rightarrow \infty} \zeta_f (t) = 0$
\item belong to a Reproducing Kernel Hilbert Space, which here is used as a tool for embedding similarity among high-dimensional and complex distributions onto lower-dimensional ones.
\end{enumerate}
The method proceeds by taking the usual expression for the loglikelihood function
\begin{align}
llh_{\tilde{T}} (\boldsymbol{\lambda}) = -\sum_{d=1}^{D} \left(\int_0^{\tilde{T}} \lambda_d (s) d s - y_{d,k} \log \lambda_d (t_k)\right),
\end{align}
and optimizing, instead, over a discretized version of it
\begin{align}
llh_{\tilde{T}} (\boldsymbol{\lambda}) &= \sum_{d=1}^D \sum_{k=1}^{M (t)} \left( \int_{\chi_{k-1}}^{\chi_k} \lambda_d (s) d s - y_{d,k} \log \lambda_i (t_k)\right) \\ &= \sum_{d=1}^D \Delta L_{d,t} (\lambda_d) ,
\end{align}
with $(t_1, ..., t_{n (t)})$ denoting the event arrival times over an interval $[0,\tilde{T}]$, and with a partitioning $\{0,\chi_1, ..., \chi_{M (t)} \}$ of this interval $[0,\tilde{T}]$ such that
\begin{equation}
\chi_{k+1} = \min_{t_i \geq \chi_k} \{ \iota*\floor{\chi_k/\iota} + \iota, t_i \},
\end{equation}
for some small $\iota > 0$. The discretized version can then be expressed as
\begin{align}
llh_{\tilde{T}}^{(\iota)} (\boldsymbol{\lambda}) &= \sum_{d=1}^D \sum_{k=1}^{M (t)} \left( \int_{\chi_{k-1}}^{\chi_k} (\chi_k - \chi_{k-1})\lambda_d (\chi_k) \right.\\ & \left.- y_{d,k} \log \lambda_i (\chi_k)\right) = \sum_{d=1}^D \Delta L_{d,\tilde{T}}^{( \iota )} (\lambda_d)
\end{align}
The optimization procedure is done at each slot of the $M (t)$ partition, taking into account:
\begin{itemize}
\item A truncation over the intensity function effect, i.e.,
\begin{equation}
\phi (t) = 0, \forall t > t_{max},
\end{equation}
so as to simplify the optimization of the integral portion of the loss. The error over this approximation is shown to be bounded by the decreasing tail assumption.
\item A Tikhonov regularization over the coefficients $\mu_d$ and $\phi_{d,d}$, which is simply the addition of weighted $||\mu_d||^2$ and $||\phi_{d,d}||^2$ terms to the loss function, so as to keep their resulting values small;
\item A projection step for the triggering function optimization part, so as to keep them all positive.
\end{itemize}
The recent improvements over frequentist nonparametric HP estimation were shown to focus on two main strategies:
\begin{itemize}
    \item Speeding up inference through replacement of the excitation matrix as objective by the matrix of cumulants, which were shown to be enough to capture the mutual influence among each pair of nodes;
    \item And an online learning procedure, which used some assumptions over the kernels (positive, decreasing tail RKHS) so as to recover estimates of the HP parameters over a single pass on some partitioning of the event arrival timeline.
\end{itemize}

\begin{table}
\centering
\begin{tabular}{c c} 
\hline
Method & Total Complexity\\
 \hline\hline
  ODE HP \cite{KZ13} & $\mathcal{O} (Iter * \tilde{D}_{\phi} (n_{max}^3 D^2 + M*(n_{max} D + n_{max}^2)))$  \\ 
 Granger Causality HP\cite{HX16} & $\mathcal{O} (Iter * \tilde{D}_{\phi} n_{max}^3 D^2)$  \\
 Wiener-Hopf Eq. HP \cite{EB16} & $\mathcal{O} (n_{max} D^2 M + D^4 M^3)$ \\
 NPHC \cite{MA17} & $\mathcal{O} (n_{max} D^2 + Iter *D^3)$ \\
 Online Learning HP \cite{YY17} & $\mathcal{O} (Iter * D^2)$  \\ 
 \hline
\end{tabular}
\caption{Comparison of Computational Complexity of parametric and nonparametric HP estimation methods, extracted from \cite{MA17}. \textit{Iter} is the number of iterations of the optimization procedure, $\tilde{D}_{\phi}$ is the number of composing basis kernels of $\phi (t)$, D is the dimensionality of the MHP, $n_{max}$ is the maximum number of events per sequence, and M is the number of components of the discretization applied to $\phi (t)$. Complexities obtained from \cite{MA17} and \cite{YY17}.}
\label{tab: comparison_complexity}
\end{table}

\begin{table} 
\centering 
\resizebox{\textwidth}{!}{
\begin{tabular}{l c c c c} 
\toprule 
& \multicolumn{4}{c}{\textbf{MHP Estimation Method}} \\ 
\cmidrule(l){2-5} 
\textbf{Performance Metric} & ODE HP \cite{KZ13} & Granger Causality HP \cite{HX16} & ADM4 \cite{KZ132} & NPHC \cite{MA17}\\ 
\midrule 
Relative Error & 0.162 & 0.19 & 0.092 & 0.071\\ 
Estimation Time (s)& 2944 & 2780 & 2217 & 38 \\ 
\midrule 
\midrule 
\end{tabular}}
\caption{Performance comparison of several Multivariate HP Estimation methods in the MemeTracker \cite{JL14} dataset, extracted from \cite{MA17}. The relative error between a ground truth impact matrix $\boldsymbol{\alpha} = \{\alpha_{ij}\}$ and its estimate $\hat{\alpha} = \{\hat{\alpha}_{ij}\}$ is simply $\sum_{i,j} |\alpha_{ij} -\hat{\alpha}_{ij}|/|\alpha_{ij}| \mathbb{1}_{\{\alpha_{ij} \neq 0\}} + |\hat{\alpha}_{ij}| \mathbb{1}_{\{\alpha_{ij} = 0\}}$.} 
\label{tab: comparison_par_nonpar} 
\end{table}

A comparison among the complexity of several parametric and frequentist nonparametric HP estimation methods is shown in Table \ref{tab: comparison_complexity}, and performance metrics are shown in Table \ref{tab: comparison_par_nonpar}. In general, it is possible to see a focus of more recent methods, such as NPHC and the Online Learning approach, in reducing the complexity per iteration of the resulting estimation procedure through approximation assumptions on the underlying model.

\subsection{Bayesian Nonparametric HPs}

Another nonparametric treatment of HPs revolves around the assumption of the triggering kernel and the background rates to be modeled by distributions (or mixtures of distributions) from the so-called ``Exponential Family'', which, through their conjugacy relationships, allow for closed-form computations of the sequential updates in the model. These were mainly proposed in the works of \cite{ND15}, \cite{SD18} and \cite{RZ19}.

In \cite{ND15}, HPs have also been used for modeling clustering of documents streams, capturing the dynamics of arrival time patterns, for being used together with textual content-based clustering.

The logic is that news and other media-related information sources revolving around a given occurrence- such as a natural catastrophe, a political action, or a celebrity scandal- are related not only regarding its word content, but also their time occurrences, as journalists tend to release more and more contents about a topic of high public interest, but tend to slow down the pace of publication as this interest gradually vanishes or shifts toward other subjects.

The main idea is to unite both Bayesian Nonparametric Inference, which is a scalable clustering method which allows for new clusters to be added as the number of samples grow, with Hawkes Processes. The corresponding Bayesian Nonparametric model, the Dirichlet Process, would capture the diversity of event types, while the HPs would capture the temporal dynamics of the event streams.

A Dirichlet Process $DP(\alpha,G_0)$ can be roughly described as a probability distribution over probability distributions. It is defined by a concentration parameter $\alpha$, proportional to the level of discretization ('number of bins') of the underlying sampled distribution, and a base distribution $G_0$, which is the distribution to be discretized. As an example, for $\alpha$ equal to 0, the distribution is completely concentrated at a single value, while, in the limit that $\alpha$ goes to infinity, the sampled distribution becomes continuous.

The corresponding hybrid model, the Dirichlet-Hawkes Process (DHP) is defined by:
\begin{itemize}
\item $\mu$, an intensity parameter;
\item $\mathbb{P}_0^{DHP} (\theta_{DHP})$, a base distribution over a given parameter space $\theta_{DHP} \in \Theta_{DHP}$;
\item A collection of excitation functions $\phi_{\theta_{DHP}} (t,t')$.
\end{itemize}

After an initial time event $t_1$ and an excitation function parameter $\theta_{DHP}^1$ are sampled from these base parameters $\mu$ and $\mathbb{P}_0^{DHP} (\theta_{DHP})$, respectively, the DHP is then allowed to alternate between:
\begin{enumerate}
\item Sampling new arrival events $t_i$ from the current value of $\theta_{DHP}$ for the excitation function, with probability 
\begin{equation}
\label{eq:em_1}
\dfrac{\mu}{\mu + \sum_{i=1}^{n-1} \phi_{\theta_{DHP}^i} (t_n,t_i)}
\end{equation}
\item Or sampling a new value for $\theta_{DHP}$ with probability 
\begin{equation}
\label{eq:em_2}
\dfrac{\sum_{i=1}^{n-1} \phi_{\theta_{DHP}^i} (t_n,t_i)}{\mu + \sum_{i=1}^{n-1} \phi_{\theta_{DHP}^i} (t_n,t_i)}
\end{equation}
\end{enumerate}
In this way, you are dealing with a superposition of HPs, in which the arrival events tend towards processes with higher intensities, i.e., the preferential attachment, but which also allows for diversity, since there is always a nonzero probability of sampling a new HP from the baseline intensity $\mu$.

By defining the excitation functions $\phi_\theta$ as a summation of parametric kernels
\begin{equation}
\phi_{\theta_{DHP}} (t_i,t_j) = \sum_{l=1}^{K} \alpha_{\theta_{DHP}}^l \kappa_{DHP}^l (t_i - t_j)
\end{equation}
the model can be made even more general.

The approach in \cite{SD18}, besides the random histogram assumption for the triggering kernel, similar to the frequentist case, also considers the case of it being defined by a mixture of Beta distributions, and have the model being updated through a sampling procedure (Markov chain Monte Carlo).

The work in \cite{RZ19} proposes a Gamma distribution over the possible values of $\mu$ and the triggering kernel to be modeled as:
\begin{equation}
\phi(\cdot) = \dfrac{\mathcal{GP} (\cdot)^2}{2},
\end{equation}
where $\mathcal{GP} (\cdot)$ is a Gaussian Process \cite{CR06}.

These assumptions allow for closed-form updates over the posterior distributions over the background rate and the triggering kernel, which are claimed to be more scalable and efficient than the plain binning of the events.

In the next section, we will explore the neural architectures, which were introduced for modeling and generation of HPs through a more flexible representation of the effect of past events in the intensity function.

\section{Neural Network-based HPs}
\label{sec: nn-based}

In this section, we discuss the neural network-based formulations of HP modeling. The main idea is to capture the influence of past events on the intensity function in a nonlinear, and thus hopefully more flexible, way. 

This modeling approach makes use of recurrent models, which, in their simplest formulation, encode sequences of states 
\begin{equation}
(z_0^s,z_1^s,...,z_N^s)
\end{equation}
and outputs
\begin{equation}
(z_0^o, z_1^o, ... , z_N^o)
\end{equation}
in a way that each state $z_{i+1}^s$ can be obtained by a composition of the immediately preceding state $z_i^s$ and a so-called hidden state $ h_i$ which captures the effect of the other past states:
\begin{equation}
h_{i} = \sigma_h (W_s  z_i^s + W_h  h_{i-1} + b_h)
\end{equation}
\begin{equation}
z_i^o = \sigma_o (W_o  z_i^o + b_o),
\end{equation}
in which $W_o $, $W_s $, $W_h $, $b_h$ and $b_o$ are parameters to be fitted by the optimization procedure, while $\sigma_h$ and $\sigma_o$ are nonlinearities, such as a sigmoid or a hyperbolic tangent function.

In the case of HPs, the state to be modeled would be the intensity function along a sequence of time event arrivals, and an additional assumption would be that its intensity value decays exponentially between consecutive events.

In the case of most NN-based models, the inference also counts with a mark distribution for the case of marked HPs, in which a multinomial, or some other multi-class distribution, is fitted together with the recurrent intensity model.

Arguably the first from such type of models, the Recurrent Marked Temporal Point Process \cite{ND16} jointly models marked event data by using a RNN with exponentiated output for modeling the intensity. 

For sequences of the type $\{t_i,y_i\}_{i=1}^{N}$, in which $t_i$ corresponds to the time of the i-th event arrival, while $z_i^{o}$ refers to the type of event or mark, we have a hidden cell $h_{i}$ described by:
\begin{equation}
\boldsymbol{ h_i} = \max \left\{ W_o  z_i^o + W^t t_i + W_h  h_{i-1} + b_h, 0\right\},
\end{equation}
and a Conditional Intensity Function defined as a function of this hidden state
\begin{equation}
\lambda (t) = exp \left(\boldsymbol{v^t} \boldsymbol{ h_i} + w_t  (t- t_i) + b^t\right),
\end{equation}
while a K-sized mark set can have its probability modeled by a Softmax distribution:
\begin{equation}
P (z_{i+1}^o = k | \boldsymbol{ h_i}) = Softmax(k,\boldsymbol{V_k^{zo}} \boldsymbol{ h_i} + b_k^{zo}) = \dfrac{exp \left(\boldsymbol{V_k^{zo}} \boldsymbol{ h_i} + b_k^{zo} \right)}{\sum_{k=1}^{K} exp \left(\boldsymbol{V_k^{zo}} \boldsymbol{h_i} + b_k^{zo}\right)}
\end{equation}
As the likelihood of the whole sequence can be defined as a product of conditional density functions for each event:
\begin{equation}
llh \left( \{t_i,z_i^o\}_{i=1}^{N} \right) = \prod_{i=1}^N f (t_i, z_i^o),
\end{equation}
with 
\begin{align}
f_\lambda (t) &= \lambda (t) exp \left(-\int_{t_n}^t \lambda (t)dt \right),\nonumber\\
\end{align}
where $t_n$ is the latest event ocurred before time t. From this $f (t)$, we may estimate the time of the next event as:
\begin{equation}
t_{i+1} = \int_{t_i}^\infty t f_\lambda (t) dt
\end{equation}
This allows us to optimize the parameters of the RNN model over the loss equal to this likelihood function, composed by these conditional density functions.

Given a set of M training sequences $\mathcal{S}^j = \{t_i^j,y_i^j\}_{i=1}^{N^j}$, we want to optimize the weight parameters over a loss function defined as:
\begin{equation}
llh (\left\{\mathcal{S}^j\right\}_{j=1}^M) = \sum_j^M \sum_i^{N_j} \log \left( P (z_{i+1}^{o,j} | \boldsymbol{ h_i}) + \log f_\lambda (t_{i+1}^j - t_i^j | \boldsymbol{ h_i})\right)
\end{equation}
This optimization procedure is usually done through the Backpropagation Through Time (BPTT) algorithm, which proceeds by `unrolling' the RNN cells by a fixed number of steps, then calculating the cumulative loss along all these steps, together with the gradients over each of W's, w's, v's and b's, then updating these parameters with a pre-defined learning rate until convergence.

One improvement for the RNN-based modeling approach is described in \cite{SX173}, referred to as ``Time Series Event Sequence'' (TSES), which consists of treating the mark sequences as derived from another RNN model, instead of the multinomial distribution. This RNN for the marks is then jointly trained with the RNN for event arrival times.

Another improvement over this NN-based modeling approach is the Neural Hawkes Process \cite{HM17}, which uses a variant of the basic RNN, entitled Long Short-Term Memory (LSTM) \cite{SH97}, applied to the intensity function modeling.

The Neural Hawkes Process models the intensity function of a multi-type event sequence by associating each k-th event type with a corresponding intensity function $\lambda_{ k} (t)$, s.t.:
\begin{equation}
\lambda_{ k} (t)=  f_k (w_{ k}^T \boldsymbol{z^h} (t)),
\end{equation}
with
\begin{equation}
\boldsymbol{z^h} (t) = \boldsymbol{o_i } \odot (2 g  (2 \boldsymbol{c }(t) - 1)),
\end{equation}
The variables $\boldsymbol{o_i }$ and $\boldsymbol{c } (t)$ are defined through the following update rules:
\begin{equation}
\boldsymbol{c_{i+1} } = f_{i+1}  \odot \boldsymbol{c } (t_i) + \boldsymbol{i}_{i+1}  \odot \boldsymbol{z}_{i+1} ,
\end{equation}
The variables 
$\boldsymbol{i}_{i+1}$,$\boldsymbol{f}_{i+1}$, $\boldsymbol{z}_{i+1}$ and $\boldsymbol{o}_{i+1}$ are defined similarly to the gated variables of a standard LSTM cell. The reader is invited to read the original paper for their full definition. And the value of the $\boldsymbol{c } (t)$ is assumed to decay exponentially among consecutive events as:
\begin{align}
\boldsymbol{c } (t) &= \boldsymbol{\overline{c}_{i+1} } + (\boldsymbol{c}_{i+1}  - \boldsymbol{\overline{c}_{i+1} }) exp (-\boldsymbol{\delta}_{i+1}  (t - t_i)),
\end{align}
for
\begin{equation}
\boldsymbol{\overline{c}_{i+1} } = \overline{f}_{i+1}  \odot \boldsymbol{\overline{c} } (t_i) + \boldsymbol{\overline{i} }_{i+1} \odot \boldsymbol{z}_{i+1} 
\end{equation}
\begin{equation}
\boldsymbol{\delta}_{i+1}  = g  (\boldsymbol{W_\delta } \boldsymbol{k_i} + \boldsymbol{U_\delta } \boldsymbol{h} (t_i) + \boldsymbol{d_\delta }).
\end{equation}
The $W $'s, $U $'s and $d $'s of the model are trained so as to maximize the loglikelihood over a set of sequences. Compared with the previous RNN-based model, in the Neural Hawkes Process:
\begin{enumerate}
\item The baseline intensity $\mu_k$ is not implicitly considered constant, but instead is allowed to vary;
\item The variations of the cell intensity are not necessarily monotonic, because the influences of each event type on the cell values may decay at a different rate;
\item The sigmoid functions along the composition equations allow for an enriched behaviour of the intensity values.
\end{enumerate}
All this contributes to an increased expressiveness of the model. Besides, as in the regular LSTM models, the ``forget'' gates $\boldsymbol{f}_{i+1}$ are trained so as to control how much influence the past values of $\boldsymbol{c} (t)$ will have on its present value, thus allowing the model to possess a ``long-term'' memory.

Another variant of the RNN-based HPs, introduced in \cite{SX17}, models the baseline rate and the history influence as separate RNNs each. The baseline rate is taken as a time series, with its corresponding RNN updating its state at equally spaced intervals, such as five days. The event history influence RNN updates its state at each event arrival. This has been shown to increase the time and mark prediction performance, as demonstrated in Table \ref{tab: nn_comparison}.

Both the background rate time series $\{\mu (t)\}_{t=1}^{T}$ and the marked event sequence $\{m_i,t_i\}_{i=1}^N$ are modeled by LSTM cells:
\begin{equation}
(\boldsymbol{h}^\mu (t),\boldsymbol{z}_c^\mu (t)) = \text{LSTM}_\mu (\boldsymbol{\mu} (t),\boldsymbol{h}^\mu (t-1)+ \boldsymbol{z}_{c}^\mu (t-1)),
\end{equation}
\begin{equation}
(\boldsymbol{h}^m (i),\boldsymbol{z}_c^m (i)) = \text{LSTM}_m (\boldsymbol{m}_i,\boldsymbol{h}^m (i)+ \boldsymbol{z}_{c}^m (i-1)),
\end{equation}
These h and c states correspond to the hidden state and the long-term dependency terms, respectively, similarly to the Neural Hawkes Process. Both terms are concatenated in a single  variable $\boldsymbol{z}_e (t)$, for jointly training both RNN models:
\begin{equation}
\boldsymbol{z}_e (t) = tanh (\boldsymbol{W}_f  \left[\boldsymbol{h}^{\mu} (t), \boldsymbol{h}^m\right]+ \boldsymbol{b}_f )
\end{equation}
\begin{equation}
\boldsymbol{U}  (t)= Softmax (\boldsymbol{W}_U  \boldsymbol{z}_e (t) + \boldsymbol{b}_U ),
\end{equation}
\begin{equation}
\boldsymbol{u}  (t)= Softmax (\boldsymbol{W}_u  \left[\boldsymbol{z}_e (t),\boldsymbol{U} (t) \right] + \boldsymbol{b}_u )
\end{equation}
\begin{equation}
z_s  = \boldsymbol{W}_s  \boldsymbol{z}_e (t) + b_s ,
\end{equation}
with $U $ and $u $ denoting the main event types and subtypes, respectively, and $z_s $ denoting the composed timestamp of each event.
The loss over which the model is trained is defined in a cross-entropy way:
\begin{align}
&\sum_{j=1}^N \left(-\boldsymbol{W}_{U}  (j)\log (U  (t,j)) - w_{u}  (j)\log (u  (t,j)) \right.\\ &\nonumber \left.- \log \left(f (z_{s}  (t,j)| h  (t-1,j))\right)\right),
\end{align}
with
\begin{equation}
f \left(z_{s}  (t,j) | h  (t-1,j)\right) = \dfrac{1}{\sqrt{2 \pi \sigma}} \mathrm{e}^{\left(-\dfrac{(z_{s}  (t,j) - z_{\tilde{s}}  (t,j))^2}{2 \sigma^2}\right)},
\end{equation}
where $z_{\tilde{s}}  (t,j)$ is the model predicted output for the corresponding event $z_{s}  (t,j)$.

The model weights are then jointly trained, over the total loss function and under some correction for the frequency  ratio of each event type, for both the background rate time series values and the event arrivals RNN, and are shown to outperform more `rigid' models.

Now, regarding the generation of HP sequences, both RMTPP and Neural Hawkes Processes have in common the fact that they intend to model the intensity function of underlying process, so that new sequences may be sampled in a way to reproduce the behaviour of the original dataset. This intensity modeling, however, has three main drawbacks:
\begin{enumerate}
\item It may be unnecessary, since the sequences may be simply produced by unrolling cells of corresponding RNN models;
\item The sequences from these intensity-modeling approaches are sampled using a `Thinning algorithm' \cite{YO81}, which may incur in slowed-down simulations, in the case of repeatedly rejected event intervals.
\item These methods are trained by maximizing the loglikelihood over the training sequences, which is asymptotically equivalent to minimizing the KL-Divergence over original and model distributions. This MLE approach is not robust in the case of multimodal distributions
\end{enumerate}
The model in \cite{SX172} proposes approximating a generative model for generating event sequences by using an alternative metric of difference among distributions, the Wasserstein (or Earth-Moving) distance, already discussed in Section \ref{sec: param}.
 
In the model, entitled Wasserstein Generative Adversarial Temporal Point Process (WGANTPP), this Wasserstein loss is shown to be equal to:
\begin{align}
L^{'} &= \left[\dfrac{1}{m} \sum_{i=1}^m F_w (G_\Theta (\mathcal{S}_s^i)) - \dfrac{1}{m} \sum_{i=1}^m F_w (\mathcal{S}_r^i)\right] \\ &+ \nu \sum_{i,j = 1}^m \dfrac{|f_w (\mathcal{S}_r^i) - F_w (G_\Theta (\mathcal{S}_r^j))|}{|\mathcal{S}_r^i - G_\Theta (\mathcal{S}_s^j)|_*},
\end{align}
where the second term, along with the constant $\nu$ correspond to the so-called Lipschitz constraints, related to the continuity of the models. 

$\{\mathcal{S}_r^i\}_{i=1}^m$ are real data sequences, while $\{\mathcal{S}_s^i\}_{i=1}^n$ are sequences sampled from a Homogeneous Poisson Process with a rate $\lambda_{HPP}$ which is simply the expected arrival rate over all training sequences. The generator network $G_\Theta$ and  discriminator $F_w$ are defined as:
\begin{equation}
G_\Theta (\mathcal{S}_r) = \tilde{\mathcal{S}} = \{t_1, ..., t_n\}
\end{equation}
with
\begin{equation}
t (i) = g_G^x (f_G^x (h(i))) \text{ and } h  (i) = g_G^h (f_G^h (z_s, h(i-1)))
\end{equation}
\begin{equation}
F_w (\tilde{\mathcal{S}}) = \sum_{i=1}^n a (i)
\end{equation}
with
\begin{equation}
a (i) = g_D^a (f_D^a(h(i)) \text{ and } h  (i) = g_D^h (f_D^h (t_i, h(i-1)))
\end{equation}
with the g's defined as nonlinearities, and $\tilde{\mathcal{S}}$ as some example time event sequence. The f's are linear transformations, as in a standard RNN cell, with their corresponding weight matrices and bias vectors to be tuned by a Stochastic Gradient Descent procedure.

\begin{figure}
    \centering
    \includegraphics[width=0.6\linewidth]{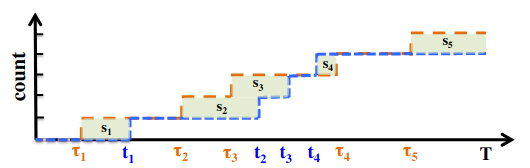}
    \caption{Intuition behind the distance metric $|\cdot|_*$ among two given event sequences $\{t_i\}$ and $\{\tau_i\}$. Extracted from \cite{SX172}.}
    \label{fig:distance_metric}
\end{figure}
The distance metric $|\cdot|_*$ of two sequences $\{t_i\}$ and $\{\tau_i\}$, for the case of purely temporal point processes in $[0,T)$, can be shown to be equivalent to
\begin{equation}
\sum_{i=1}^n |t_i - \tau_i| + (m-n) \times T - \sum_{i=n+1}^m \tau_i,
\end{equation}
which has an intuitive graphical interpretation, as shown in Figure \ref{fig:distance_metric}.

As previously discussed, the generator is trained so as to ``fool'' the discriminator, while the discriminator is trained so as to distinguish generated sequences from those of real data. This adversarial training procedure is roughly equivalent to gradient updates with opposite signs over their respective parameters: positive sign for the discriminator and negative sign for the generator. 

In the end of the training procedure, one hopefully gets a generator network capable of producing sequences virtually indistinguishable from the real data ones.

This method, however, consists of training a network for generating entire sequences, and so the generator model learned may not accurately generate conditional output sequences from input ones. Another model, described in \cite{SX18}, deals with this task by generating, from partially observed sequences, the future events of these same sequences conditioned on their history, i.e., instead of aiming to capture the underlying distribution of a set of full sequences, the model performs a "sample agnostic" in-sample prediction.

Analogously to one of the parametric models described in Section \ref{sec: param}, the learning procedure of this in-sample neural-network based prediction model takes advantages of both types of divergence measures: MLE loss (or KL Divergence) and Wasserstein distance. 

The former aims for a rigid and unbiased parameter matching among two given probabilistic distributions, which is sensitive to noisy samples and outliers,  while the latter has biased parameter updates, but is sensitive to underlying geometrical discrepancies among sample distributions. This combined loss is a way to balance both sets of priorities. In the case of long-term predictions, in which initial prediction errors propagate and magnify themselves throughout the whole stream, this joint loss was found to strengthen the effectiveness of the inference procedure.

The proposed model borrows on the seq2seq architecture (\cite{IS14}) and aims to model endings of individual sequences conditioned on their partially observed history of initial events, and inserts an adversarial component in the training to increase the accuracy of long-term predictions. A network, designated as generator, encodes a compact representation of the initial partial observation of the sequence and outputs a decoded remaining of this same sequence. That is to say that, for a full sequence:
\begin{equation}
\{t_1, t_2, ..., t_{n+m} \},
\end{equation}
the seq2seq modeling approach learns a mapping
\begin{equation}
G_\Theta (\mathcal{S}^{1,n}) = \mathcal{S}^{m,n}
\end{equation}
such that
\begin{equation}
\mathcal{S}^{1,n} = \{ t_1, t_2, ..., t_n\} \text{ and } \mathcal{S}^{m,n} = \{t_{n+1}, t_{n+2}, ..., t_{n+m}\}.
\end{equation}


This mapping is defined through a composition of RNN cells:
\begin{equation}
\boldsymbol{h}_i = \eta_g^h (f_A^h (t_i,\boldsymbol{h}_{i-1}) \text{ and } t_{i+1} = \eta_g^x f_g^x (\boldsymbol{h}_i) 
\end{equation}

with the $\eta$'s defined as nonlinear activation functions, and the f's as linear transformations with trainable weight matrices and bias vectors.

The learning procedure consists of tuning the RNN cells' parameters so as to maximize the conditional probability
\begin{equation}
P (\mathcal{S}^{m,n} | \mathcal{S}^{1,n}) = \prod_{i=n}^{n+m-1} P (t_{i+1} | h_i, t_1, ..., t_i),
\end{equation}
what is carried through by parameter gradient updates over the combined loss Wasserstein and MLE loss. The adversarial component of the training, the discriminator $F_w^{S2S} (\cdot)$, is modeled as a residual convolutional network (see \cite{KH16}) with a 1-Lipschitz constraint, which is related to the magnitude of the gradients of the discriminative model. The full optimization problem then becomes 
\begin{align}
&\min_\theta \max_w \underbrace{\sum_{l=1}^M F_w^{S2S} (\{\mathcal{S}_{l}^{1,n},\mathcal{S}_{l}^{m,n}\}) - \sum_{l=1}^{M} F_w^{S2S} (\{\mathcal{S}_{l}^{1,n},G_\Theta (\mathcal{S}_l^{1,n}) \})}_{\text{Wasserstein loss}} \\ &- \underbrace{\gamma_{LIP} | \frac{\partial F_{w'}^{S2S} (\hat{x})}{\partial \hat{x}} - 1|}_{\text{1-Lipschitz constraint}} - \underbrace{\gamma_{MLE} log (\mathbb{P}_\theta (\mathcal{S}^{m,n} | \mathcal{S}^{1,n}))}_{\text{MLE loss}}
\end{align}

Further works on RNN-based modeling of HPs can also be found in \cite{TO19} and \cite{SJ20}.

\subsection{Self-Attentive and Transformer Models}

Another improvement for NN-based modeling, proposed in \cite{QZ19}, involves a so-called self-attention strategy \cite{AV17} to improve the accuracy of the resulting network. The i-th event tuple $(t_i, m_i)$ is embedded as a variable $x_i$:
\begin{equation}
\boldsymbol{z_i} = \boldsymbol{tp_{m}} + \boldsymbol{pe_{(m_i,t_i)}},
\end{equation}
which simultaneously encodes information about the event mark through
\begin{equation}
\boldsymbol{tp_{m}} = \boldsymbol{z_e^m} \mathbb{W}_E,
\end{equation}
with $\boldsymbol{z_e^m}$ as an one-hot encoding vector of the mark and $\mathbb{W}_E$ as an embedding matrix, and information about the time interval among consecutive events through a sinusoidal-based positional encoding vector $\boldsymbol{pe_{(m_i,t_i)}}$, with its k-th entry defined as:
\begin{equation}
pe^k_{(m_i,t_i)} = sin(\omega_k^i \times i + \omega_k^t \times t_i)
\end{equation}
From this encoded variable $\boldsymbol{x_i}$, a hidden state $\boldsymbol{h_{u,i}}$ is then defined for each category $u$ of the marks, which captures the influence of all previous events:
\begin{equation}
\boldsymbol{h_{u,i+1}} = \dfrac{\left( \sum_{j=1}^{i} f(\boldsymbol{z_{i+1},\boldsymbol{z_j}}) \boldsymbol{g(z_j)}\right)}{ \sum_{j=1}^{i} f(\boldsymbol{z_{i+1},\boldsymbol{z_j}})}
\end{equation}

\begin{table} 
\centering 
\resizebox{\textwidth}{!}{
\begin{tabular}{c c c c c c || c c c || c c c } 
\toprule 
& \multicolumn{5}{c}{\textbf{Loglikelihood per Event}} & \multicolumn{3}{c}{\textbf{Time Prediction RMSE}} & \multicolumn{3}{c}{\textbf{Mark Prediction Accuracy}}\\
\midrule
\textbf{Method} $\backslash$ \textbf{Dataset} & RT & MT & FIN & MIMIC-II & SO & FIN & MIMIC-II & SO & FIN & MIMIC-II & SO\\ 
\midrule 
RMTPP \cite{ND16}& -5.99 & -6.04 & -3.89 & -1.35 & -2.60 & 1.56 & 6.12 & 9.78 & 61.95 & 81.2 & 45.9\\ 
Neural HP \cite{HM17}& -5.60 & -6.23 & -3.60 & -1.38 & -2.55 & 1.56 & 6.13 & 9.83 & 62.20 & 83.2 & 46.3\\ 
TSES \cite{SX17} & - & - & - & - & - & 1.50 & 4.70 & 8.00 & 62.17 & 83.0 & 46.2\\ 
Self-Attentive HP \cite{QZ19} & -4.56 & - & - & -0.52 & -1.86 & - & 3.89 & 5.57 & - & - & -\\ 
Transformer HP \cite{SZ20} & -2.04 & 0.68 & -1.11 & 0.82 & 0.04 & 0.93 & 0.82 & 4.99 & 62.64 & 85.3 & 47.0\\ 
\midrule 
\bottomrule 
\end{tabular}}
\caption{Performance comparison of NN-based HP models, regarding: a) Loglikelihood averaged per number of events; b) RMSE of predicted time interval; and c) Accuracy of mark prediction. The performances were measured over sequences from Retweet \cite{QZ15}, MemeTracker \cite{JL14}, Financial \cite{ND16}, Medical Records \cite{AJ16}, and Stack Overflow \cite{JL14} datasets. The TSES method is a likelihood-free model, and so its entries are not evaluated for the \textbf{Loglikelihood per Event} section of the table. Values are obtained from \cite{SZ20}} 
\label{tab: nn_comparison} 
\end{table}

Through a series of nonlinear transformations, the intensity $\lambda_u (t)$ for the u-th mark is then computed. A concurrently developed approach in \cite{SZ20} uses multiple Attention layers to build a so-called ``Transformer Hawkes Process'', which also surpasses the performance of RNN-based approaches in a series of datasets, as shown in Table \ref{tab: nn_comparison}

\subsection{Graph Convolutional Networks}

A further improvement of neural HP models, described in \cite{JS19}, involves the graph properties of multivariate HPs, which may be embedded in a NN modeling framework through the recently proposed Graph Convolutional Networks (GCN)\cite{TK16}. 

The method is composed of a GCN module, for capturing meaningful correlation patterns in a large sets of event sequence, followed by an usual RNN module, for modeling the temporal dynamics. In short, the time sequences are
modeled as a HPs, and the adjacencies among
different processes are encoded as a graph. The novel (GCN+RNN) model is meant to extract significant local patterns from the graph. 
The output of the initial GCN network module, which is simply a matrix of the form $\boldsymbol{\chi} = [\boldsymbol{\mu},\mathbb{A}]$, which includes the baseline vector $\boldsymbol{\mu}$ and the adjacency matrix $\mathbb{A}$, are fed into the RNN-based module, there taken as a LSTM. Then, the output of this RNN module are input to a further module, a fully connected layer, for calculating
the changes $d\mathbb{X}$ to be applied to the current parameter matrix $\mathbb{X}$. Then, after each training step T, the predicted value of this parameter matrix becomes as $$\mathbb{X} (T) = \mathbb{X} (T-1) + d \mathbb{X} (T-1)$$.

The work in \cite{GY18} proposes a model for check-in time prediction which is composed by a LSTM-based module, in which the feature vector is composed so as to capture each relevant aspect of the problem: the event time coordinate $t_i$, an additional field to indicate if the check-in occurred on a weekday or during the weekend, the euclidean distance between a given check-in location and the location (l) of the previous check-in, the location type of the check-in (e.g., Hotel, Restaurant, etc.), the number of users overlapping with a given location, and the check-ins by friends of the user. This aggregates social, geographical and temporal information in a single NN-based HP-like predictive model.

All these variants of neural-network point process models allow for more flexible (nonlinear) representation of the effect of past events in the future ones, besides putting at the inference procedure's disposal a myriad of Deep Learning tools and techniques which have enjoyed a surge of popularity over the recent years.

\section{Further Approaches}
\label{sec: further}

In this section, we briefly review some recently proposed approaches which do not fit conveniently into any of the three previously discussed subgroups, but may be considered as bridgings between usual HP-related tasks and other mathematical subfields.

\subsection{Sparse Gaussian Processes}

By building over the modeling in \cite{RZ19}, which considers the triggering function of the HP as a Gaussian Process, the work in \cite{RZ192} proposes an approach involving Sparse Gaussian Processes for optimizing over a dataset. In this, the optimization of the likelihood is being taken not over the samples, but over a set of much fewer so-called inducing points, which are also taken as latent variables, so as to result in a final model which is both expressive enough to capture the complexity of the dataset but also tractable enough to be useful and applicable to reasonably-sized datasets.

\subsection{Stochastic Differential Equation}

Another way of modeling HPs is through a Stochastic Differential Equation, as proposed in \cite{YL16}. In them, the decay of the triggering kernel is taken as exponential, but its amplitude is defined as an stochastic process, there proposed as being either a Geometric Brownian Motion or an exponentiated version of the Langevin dynamics.

\subsection{Graph Properties}

Besides some works, previously discussed, which deal with the properties of the excitation matrices of the multivariate HPs as terms to be optimized jointly with other parameters, such as the Graph Convolutional approaches and the sparsity inducing penalization terms of some parametric and non-parametric approaches, there have been several other optimization strategies taking into account other properties of these matrices. 

\cite{SL14} explicitly inserts considerations of the excitation matrix as a distribution over some types of randomly generated matrices into the optimization of the HP likelihood. \cite{YL182} introduces a penalization term involving the proximity of the excitation matrix to a so-called connection matrix, defined to capture the underlying connectivity among the nodes of the multivariate HP, to the parameter optimization strategy. \cite{DL19} introduces a weighted sum of the Wasserstein Discrepancy and the so-called Gromov-Wasserstein Discrepancy as a penalizing factor on the usual MLE procedure of HP estimation, with the intention of inducing both absolute and relational aspects among the nodes of the HP. \cite{EB154} introduces, besides the sparsity-inducing term, another one related to the resulting rank of the excitation matrix, which is designed to induce resulting matrices which are composed of both few nonzero entries and also few independent rows.

\subsection{Epidemic HPs}

The work in \cite{MR182} blends the HP excitation effect with traditional epidemic models over populations. By considering a time event as an infection, it models the diffusion of a disease by introducing a HP intensity function which is modulated by the size of the available population, as below:
\begin{equation}
\lambda (t) = \left( 1 - \dfrac{N_t}{\tilde{N}}\right) \{\mu + \sum_{t_i < t} \phi(t-t_i)\},
\end{equation}
where $N_t$ denotes the counting process associated with the HP, while $\tilde{N}$ is the total finite population size.

\subsection{Popularity Prediction}

The work in \cite{SM16} proposes the use of HP modeling blended with other Machine Learning techniques, such as Random Forests, to obtain an associated so-called ``Popularity'' measure, which is defined as the total number of events the underlying process is expected to generate as $t \rightarrow \infty$. This measure is treated as an outcome derived from features associated with some entity (e.g., social network user), s.a. number of friends, total number of posted statuses and the account
creation time.

In the next section, we will discuss models in which one not only wants to capture the temporal dynamics, but also wishes to influence it towards a certain goal, implicitly defined through a so-called reward function.


\section{Stochastic Control and  Reinforcement Learning of HPs}

In this section, we briefly review some control strategies regarding HPs. In some cases, one may wish not only to be able to model the traces of some event sequences, or to capture the underlying distribution of said sequences, but also try to influence their temporal dynamics towards more advantageous ones. These cases are considered in the works on Stochastic Control and Reinforcement Learning approaches for HPs.

This concept of advantageous is explicited through the definition of a so-called Reward Function, which is defined in terms of specific , and sometimes application-specific, properties of the sequences. Most works related to the subject deal with Social Network applications, and one example of Reward can be the total time the post of a user stays at the top of the feed of his/her followers. One type of reward which is not domain-specific is the dissimilarity among two sets of sequences, computed through mappings such as ``kernel mean embeddings'' \cite{KM17}. In the case of Imitation Learning approaches, one still focus on solely modeling the HPs, without steering it towards desirable behaviours. In these, the reward function is simply defined over how well the samples of the chosen model to be adjusted approximates the samples of the original HP.

\subsection{Stochastic Optimal Control}

One example of this control approach is described in \cite{AZ17}, in which the variable to be controlled are the times to post of a given user, implicitly defined as an intensity function, so as to maximize the reward function $\boldsymbol{r} (t)$, here computed as the total time that this user's posts stay at the top of the feed of his/her followers.

This ``when-to-post'' problem can be formulated as:
\begin{equation}
\min_{u (t_0,t_f) } \mathbb{E}_{(N_i,M_{\i}) (t_0,t_f]} \left[ \Omega (\boldsymbol{r} (t_f)) + \int_{t_0}^{t_f} \boldsymbol{L} (\boldsymbol{r} (\tau), u (\tau)) d \tau\right]
\nonumber
\end{equation}
\begin{equation}
\text{subject to } u (t) \geq 0, \forall t \in (t_0,t_f],
\end{equation}
where:
\begin{itemize}
\item i is the index of the broadcaster of the posts;
\item $N_i (t)$ is the Counting Process of the i-th broadcaster, with $\boldsymbol{N} (t)= \{ N_i (t)\}_{i=1}^{n}$ being an array of Counting Processes along all the n users of the network;
\item $\mathbb{A} \in \{0,1\}^{n \times n}$ is the Adjacency Matrix of the network;
\item $\boldsymbol{M}_{\i} (t) = \mathbb{A}^T \boldsymbol{N} (t) - \mathbb{A}_i N_i (t)$, which means that $M_{\i} (t)$ is the sum of the Counting Process of all users connected to user i excluding user i him/herself;
\item $t_0$ and $t_f$ are, respectively, the starting and ending times of the problem horizon taken in consideration;
\item $u (t) = \mu_i (t)$, the controlled variable, is the baseline intensity of user i, to be steered towards the maximization of the reward function;
\item $\Omega (\boldsymbol{r} (t_f))$ is an arbitrarily defined penalty function;
\item $\boldsymbol{L} (\boldsymbol{r} (\tau), u (\tau))$ is a nondecreasing convex loss function defined w.r.t. the visibility of the broadcaster's posts in each of his/her followers' feeds.
\end{itemize}
The approach used in the problem is by defining an optimal cost-to-go $J (\boldsymbol{r} (t_{f}),\lambda (t), t)$:
\begin{align}
&J (\boldsymbol{r} (t_{f}),\lambda (t), t) = \min_{u (t,t_f]} \mathbb{E}_{(N,M)(t,t_f]} \left[\phi (\boldsymbol{r} (t_f)) + \int_{t}^{t_f} \boldsymbol{l} (\boldsymbol{r} (\tau), u (\tau)) d \tau\right],
\end{align}
and find the optimal solution through the Bellman's Principle of Optimality:
\begin{align}
\nonumber
&J (\boldsymbol{r} (t), \lambda (t), t) = \min_{u (t,t+dt]} \left\{\mathbb{E} [J (\boldsymbol{r} (t+dt), \lambda (t+dt), t+dt) ] + \boldsymbol{l} (\boldsymbol{r} (t), u (t)) dt \right\} 
\end{align}
For example, in the case of a broadcaster with one follower ($\boldsymbol{r} (t) = r (t)$), if the penalty and loss functions are defined as:
\begin{equation}
\phi (r (t_f)) = \dfrac{1}{2} r^2 (t_f)
\end{equation}
and
\begin{equation}
\boldsymbol{L} (r (t), u (t)) = \dfrac{1}{2} s (t) r^2 (t) + \dfrac{1}{2} q u^2 (t),
\end{equation}
for some positive significance function $s(t)$ and some trade-off parameter q, which calibrates the importance of both visibility and number of posts, we set the derivative of $J (r (t), \lambda (t), t)$ over $u (t)$ to 0 and solve it to get to an analytical solution 
\begin{equation}
u^* (t) = q^{-1} [J (r (t), \lambda (t), t)-J (0,\lambda (t),t) ],
\end{equation}
which is thus the optimal intensity a broadcaster must adopt to maximize visibility, constrained on the cost associated to the number of posts, along this one follower's feed. Further derivations are provided to the more natural and general case, in which the broadcaster may have multiple followers.

An earlier version of this type of Stochastic Optimal Control-based approach to influence activity in social networks can be found in \cite{AZ172}. In this, the goal is to maximize the total number of actions (or events) in the network. Analogously to the previously discussed algorithm, one may solve the Continuous Time version of the Bellman Equation through defining a optimal cost-to-go $J (\boldsymbol{\lambda} (t), t)$, which here depends only on the intensities of the nodes and the time.

The control input vector $\boldsymbol{u} (t)$ acts on the network by increasing the original vector of uncontrolled intensities
\begin{equation}
\boldsymbol{\lambda} (t) = \boldsymbol{\mu}_0 + \mathbb{A} \int_0^t \kappa (t-s) d\boldsymbol{N} (s)
\end{equation}
with the equivalent rates of an underlying Counting Process vector $d \boldsymbol{M} (s)$, such that the new controlled intensity vector $\boldsymbol{\lambda}^* (t)$ is now described by
\begin{equation}
\boldsymbol{\lambda}^* (t) = \boldsymbol{\mu}_0 + \mathbb{A} \int_0^t \kappa (t-s) d\boldsymbol{N} (s) + \mathbb{A} \int_0^t \kappa (t-s) d\boldsymbol{M} (s),
\end{equation}
where $\kappa (t) = e^{- \beta t}$ in the model.
 
Then, in the same way, by differentiating the equivalent $J(\boldsymbol{\lambda} (t), t)$ over the control input, setting the corresponding expression to 0, then defining
\begin{equation}
\boldsymbol{L} (\boldsymbol{\lambda} (t), \boldsymbol{u} (t)) = -\dfrac{1}{2} \boldsymbol{\lambda}^T (t) \boldsymbol{Q} \boldsymbol{\lambda} (t) + \dfrac{1}{2} \boldsymbol{u}^T (t) \boldsymbol{S} \boldsymbol{u} (t)
\end{equation}
and
\begin{equation}
\Omega (\boldsymbol{\lambda} (t_f)) = - \dfrac{1}{2} \boldsymbol{\lambda}^T (t_f) \boldsymbol{F} \boldsymbol{\lambda} (t_f),
\end{equation}
with previously defined symmetric weighting matrices $\boldsymbol{Q}$, $\boldsymbol{F}$ and $\boldsymbol{S}$, we arrive to a closed-form expression for the optimal control intensity value
\begin{equation}
\boldsymbol{u}^* (t) = -\boldsymbol{S}^{-1} \left[\mathbb{A}^T \boldsymbol{g} (t) + \mathbb{A}^T \boldsymbol{H} (t) \boldsymbol{\lambda} (t) + \dfrac{1}{2} diag (\mathbb{A}^T \boldsymbol{H} (t) \mathbb{A})\right],
\end{equation}
where $\boldsymbol{H} (t)$ and $\boldsymbol{g} (t)$ can be computed by solving the differential equations
\begin{equation}
\boldsymbol{\dot{H}} (t) = (\beta \boldsymbol{I} - \mathbb{A})^T \boldsymbol{H} (t) + \boldsymbol{H} (\beta \boldsymbol{I} - \mathbb{A}) + \boldsymbol{H} (t) \mathbb{A} \boldsymbol{S}^{-1} \mathbb{A}^T \boldsymbol{H} (t) \boldsymbol{Q}
\end{equation}
\begin{align}
\nonumber
\boldsymbol{\dot{g}} (t) &= \left[ \beta \boldsymbol{I} - \mathbb{A}^T + \boldsymbol{H} (t) \mathbb{A} \boldsymbol{S}^{-1} \mathbb{A}^T \right] \boldsymbol{g} (t) - \beta \boldsymbol{H} (t) \boldsymbol{\mu_0}  \\  &+ \dfrac{1}{2} \left[ \boldsymbol{H} (t) \mathbb{A} \boldsymbol{S^{-1}} - \boldsymbol{I}\right] diag (\mathbb{A}^T \boldsymbol{H} (t) \mathbb{A}),
\end{align}
with final conditions $\boldsymbol{g} (t_f) = 0$ and $\boldsymbol{H} (t_f) = - \boldsymbol{F}$. The solution is constant between two consecutive events and gets recomputed at each event arrival.

\subsection{Reinforcement Learning}

The described SOC-based approaches have two main drawbacks:
\begin{itemize}
\item The functional forms of the intensities and mark distributions are constrained to be from a very restricted class, which does not include the state-of-the-art RNN-based HP models, such as those described in Section \ref{sec: nn-based};
\item The objective function being optimized is also restricted to very specific classes of functions, so as to keep the tractability of the problem.
\end{itemize}
For circumventing these drawbacks, there have been some proposed approaches which combine more flexible and expressive HP models with robust stochastic optimization procedures independent from the functional form of the objective function.

On of these methods, entitled ``Deep Reinforcement Learning of Marked Temporal Point Processes'' \cite{UU18}, works by, given a set of possible actions and corresponding feedbacks, which are both expressed as temporal point processes jointly modeled by a RNN-based intensity model $\lambda_{\theta}^* (t)$:
\begin{equation}
\lambda_{\theta}^* (t) = exp(b_{\lambda} + w_t (t-t_i) + \boldsymbol{V_{\lambda}} \boldsymbol{h_i} ),
\end{equation}
with
\begin{equation}
\nonumber
\boldsymbol{h_i} = tanh (\boldsymbol{W_h} \boldsymbol{h_{i-1}} + \boldsymbol{W_1} \mathcal{T}_{i} + \boldsymbol{W_2} \boldsymbol{y_i} + \boldsymbol{W_3} \boldsymbol{z_i} \boldsymbol{W_4} \boldsymbol{b_i} + \boldsymbol{b_h}), 
\end{equation}
where
\begin{equation}
\nonumber
\mathcal{T}_{i} = f_T (t_i - t_{i-1}) \text{ and } \boldsymbol{b_i} = f_b (1-b_i,b_i) 
\end{equation}
\begin{equation}
\nonumber
\boldsymbol{y_i} = f_y (y_i) \text{, if } b_i=0 \text{ and } \boldsymbol{z_i} = f_z (z_i) \text{, if } b_i = 1.
\end{equation}
The term $b_i$ is an indicator function to whether the i-th event is an action or a feedback. By taking the weight matrices and bias vectors from all the linear transformations f a parameter vector $\theta$, what the algorithm wishes to do is to update this vector with the gradients of each parameter over an expected Reward Function $J (\theta)$:
\begin{equation}
\theta_{l+1} = \theta_l + \eta_l \nabla_\theta J(\theta)|_{\theta = \theta_l}
\end{equation}
\begin{equation}
\nabla_\theta J (\theta) = \mathbb{E}_{\mathcal{U}_T ~ p_{\mathcal{U},\theta}^* (\cdot), \mathcal{F}_T ~ p_{\mathcal{F},\phi}^* (\cdot)} [R^* (T) \nabla_\theta log \mathbb{P}_\theta (\mathcal{U}_T)],
\end{equation}
where
\begin{equation}
p_{\mathcal{U},\theta}^* = (\lambda_\theta^*, m_\theta^*)
\end{equation}
is the joint conditional intensity and mark distribution for action events, and
\begin{equation}
p_{\mathcal{F},\phi}^* = (\lambda_\phi^*,m_\phi^*)
\end{equation}
is the joint conditional intensity and mark distribution for feedback events. The Reward $R (T)$ is defined over some domain-specific metric, which may involve the responsiveness of followers in a social network setting, or the effectiveness of memorization of words in a foreign language, in a spaced-repetition learning setting.

\subsection{Imitation Learning}

Another way of using this reinforcement learning approach is through a technique entitled ``Imitation  Learning'' \cite{SL18}. The reasoning behind it is to treat the real-data sequences as generated by an expert and then, using RNN-based sequence generation, try to make these models approximate the real-data sequence as closely as possible. 

Thus, the reward function to dictate the proportion in which the gradient of the parameters over each sequence are going to be considered is equal to how much this given sequence is likely to be drawn from the underlying distribution over the real data. This similarity is computed through a Reproducing Kernel Hilbert Space (RKHS).

The theory of RKHS is very extense, and it is not our goal to give a detailed account of it here. The key idea is that, to compute similarities among items of a given space, you compute inner products between them. Taking two items, $x_1$ and $x_2$, and computing a so-called Positive Definite Kernel (PDS) $K(x_1,x_2)$ is equivalent to computing a inner product among these two items in a high-dimensional, and potentially infinite-dimensional, vector space. The PDS used in the corresponding paper is the Gaussian kernel.

The reward function is then defined as:
\begin{equation}
\hat{r}^{*} (t) \propto \dfrac{1}{L} \sum_{l=1}^L \sum_{i=1}^{N_{T}^{(l)}} \mathbb{K} (s_{i}^{(l)},l) - \\
\dfrac{1}{M} \sum_{m=1}^{M} \sum_{i=1}^{N_T^{m}} \mathbb{K}(t_{i}^{(m)},t),
\end{equation}
where:
\begin{itemize}
    \item $L$ is the number of expert trajectories;
    \item $M$ is the number of trajectories generated by the model;
    \item $\mathbb{K} (\cdot,\cdot)$ is the reproducing kernel operator;
    \item $s_{i}^{(l)}$ is the i-th time coordinate of the l-th expert trajectory;
    \item $t_{i}^{(m)}$ is the i-th time coordinate of the m-th model-generated trajectory.
\end{itemize}
The parameters of the model are then updated through a Gradient Descent-based approach towards convergence, in which the model is expected to generate sequences indistinguishable from the real-world process.

\section{Real-world Data Limitations}

One key aspect of HP modeling which inevitably encompasses all the previously mentioned approaches is that of their applicability to real-world datasets. These may present a series of sistematic issues on the training and testing sequences, which may completely hinder the generalization of the models. We discuss some key issues here, along with some recently proposed methods, designed to handle each of them.

\subsection{Synchronization Noise}

Temporal data, specially multivariate ones, may have their event streams being extracted from distributed sensor networks. A key challenge regarding them is that of synchronization noise, i.e., when each source is subject to an unknown and random time delay.

In these cases, an inference procedure which neglects these time delays may be ignoring critical causal effects of some events over others, thus resulting in a poorly generalizing model. The work in \cite{WT19} deals specifically with this aspect of real-data HP modeling, and proposes, for the exponential triggering functions HPs, including the random time shift vector (one entry for each distinct event stream) as a parameter in the HP model, which results in an inference procedure of the following form:
\begin{equation}
\boldsymbol{\hat{z}}, \boldsymbol{\hat{\theta}} = \argmax_{\boldsymbol{z} \in \mathbb{R}, \theta \geq 0} \log \mathbb{P} (\boldsymbol{\tilde{t}} | \mathcal{N}, \boldsymbol{\theta}),
\end{equation}
where $\mathcal{N}$ is the random noise vector, and $\boldsymbol{\theta}$ corresponds to the parameters of the original exponential HP model.

\subsection{Sequences with few events}

In many domains, the availability of data is scarce, and the event streams will be composed of too few events, which results in noisiness of the likelihood and, as a consequence, unreliability of the fitted HP model, which calls for strong regularization strategies over the objective functions to be optimized. 

For this type of situation, an approach, presented in \cite{FS19}, deals with HPs with triggering functions defined as exponentials and also as a mixture of gaussian kernels, as in \cite{HX16}. Then, the parameters left to search are the background rates vector $\boldsymbol{\mu}$ and the tensor of weightings $\boldsymbol{A}$ for the excitation functions. The optimization is done through an Variational Expectation-Maximization algorithm which takes the distributions over these parameters as gaussians, and optimize, through Monte Carlo sampling, over an Evidence Lower Bound (ELBO) of their corresponding loglikelihood over a set of sequences.

\subsection{Sequences with Missing data}

Another issue in HP modeling revolves around learning from incomplete sequences, i.e., streams in which one or more of the events are missing. For this type of problem, two rather distinct approaches were recently proposed:
\begin{enumerate}
\item The first one, presented in \cite{CS18}, is applied to exponential and power-law HPs, and consists of a Markov Chain Monte Carlo-based inference over a joint process implicitly defined by the product of the likelihoods of the observed events and of the so-called virtual event auxiliary variables, which are candidates for unobserved events. This virtual variable is weighted through a parameter $\kappa$, which is related to the percentage of missing events with respect to the total event count;
\item The second one, introduced in \cite{HM19}, proposes finding the missing events over the sequences through importance weighting of candidate filling event subsequences generated by a bidirectional LSTM model built on top of the Neural Hawkes Process \cite{HM17}. 
\end{enumerate}

\section{Application Examples}

In this section, we use our HP modeling background obtained so far and apply it in case studies for three different domains: Retweeting behaviour in Social Networks, Earthquake aftershocks, and COVID-19 contact tracing. We hope this will encourage the reader to consider HPs as a modeling choice for a broad scope of applications.

\subsection{Retweet}

In \cite{HD20}, HP are used jointly with Latent Dirichlet Allocation (LDA) \cite{DB03} models for distinguishing among genuine and fake (i.e., artificially induced) retweeting of posts among Twitter users.

Given a set of 2508 users, with each j-th user corresponding to a sequence 
\begin{equation}
    \mathbf{RT}^j = \{(t_i^j,\mathcal{W}_i^j)\}_{i=0}^{N_j},
\end{equation}
, where $t_i^j$ corresponds to the timestamp associated with the i-th retweet from the j-th user, and $\mathcal{W}_i^j$ as the text content of the corresponding retweet.

The 2508 corresponding sequences were manually separated between \textit{genuine} and \textit{fake} users and labelled as such, based on a set of criteria (e.g., the content in a most of said user's retweets contains spammy links and common spam keywords, multiple retweets from a given user contain promotional/irrelevant text, the user biographical information is fabricated or contained promotional activity, or a large number of tweets or retweets were posted within a very short time window of just a few seconds).

The two resulting disjoint sets were used for training two LDA models, $LDA_f$ and $LDA_g$, for modeling the topics of fake and genuine retweeters, respectively. Given a predefined number of possible topics, here set as 10, and a given text content $\mathcal{W}_i^j$, the LDA model outputs a 10-element vector, with the probabilities of the $\mathcal{W}_i^j$ corresponding to each of the 10 possible topics.

The 10-sized vector $\mathcal{V}_f$ from $LDA_f$ is concatenated with the 10-sized vector $\mathcal{V}_g$ from $LDA_g$ and, together with the baseline intensity $\mu$ and the decay $\beta$ of an exponential HP with $\phi(t) = \mathrm{e}^{-\beta t}$, fitted over the $t_i^j$'s of each j-th sequence, forms a feature vector
\begin{equation}
    \{\mathcal{V}_f^j, \mathcal{V}_g^j,\mu^j,\beta^j\},
\end{equation}
which is then fed to a clustering algorithm, which aims to correctly classify each of the 2508 retweet sequences among fake and genuine ones. The intuition behind this hybrid HP-LDA model is to use temporal features, from the HP modeling, together with context (written) ones from the users to improve the resulting detection algorithm.

\subsection{Earthquake aftershocks}

It is well known from the study of earthquake-related time series that a strong first seismic shock gives way to a series of weaker aftershocks, which occur in a very restricted time window \cite{YO99}. 

For modeling this self-exciting property of aftershocks' arrivals, \cite{YO99} proposes a power-law self-triggering kernel
\begin{equation}
\phi_{PWL} (t, \theta_{PWL}) = \dfrac{K}{(t+c)^{p}},
\end{equation}
with $\theta_{PWL} = (K,c,p) \in \mathbb{R}_{+}^3$ as trainable parameters. This model, together with an additional baseline rate parameter $\mu$, is fitted over the temporal sequence $\{t_1, t_2, ..., t_n \}$ of aftershock timestamps with a MLE optimization procedure, such as the one in Equation \ref{eq: MLE}, in which, due to the simple parametric form of the equation for the intensity, the loglikelihood can be given in closed-form.

\subsection{COVID-19}

In \cite{WC201}, a spatiotemporal HP-inspired model is used to predict the daily rates of cases and deaths associated with the Sars-CoV-2 pandemic.

Given:
\begin{itemize}
    \item A sequence $\mathcal{S} = \{t_1, t_2, ..., t_n \} \in \mathbb{Z}_{+}^n$ of timestamps (or dates);
    \item A baseline rate $\mu_c$;
    \item A probability distribution for inter-infection time, assumed to be a Weibull distribution with shape $\alpha$ and scale $\beta$;
    \item A vector $\boldsymbol{\mathcal{M}}_c^t = \{\mathcal{M}_1^t, \mathcal{M}_2^t, ...\}$ of mobility indices, measured in percentual increase/decrease with respect to standardized values for each activity category (Recreation, Groceries, Parks, etc);
    \item A vector $\boldsymbol{\mathcal{D}}_c^t = \{\mathcal{D}_1^t, \mathcal{D}_2^t, ...\}$ of static demographic features, such as percentage of smokers, population density, and number of ICU beds;
    \item A parameter $\Delta$ for capturing a potential delay between a change on mobility indices and the time $t_j$ of a primary infection being reported.
\end{itemize}

A model of the rates of cases (or deaths) $\lambda_c$ is built as
\begin{equation}
    \lambda_c = \mu_c + \sum_{t>t_j, t_j \in \mathcal{S}} \mathcal{R}_c^{t_j} (\boldsymbol{\mathcal{D}}_c, \theta_{\mathcal{D}}) \times \mathcal{R}_c^{t_j} (\mathbf{m}_c^{t_j -\Delta}, \theta_m) w(t-t_j),
\end{equation}
where $\mathcal{R}_c^{t_j}$ refers to the Reproduction Number, which is the number of people a given infected individual is expected to transmit the disease to. This $\mathcal{R}_c^{t_j}$ is model through a Poisson regression
\begin{equation}
    \mathbb{E} \left[\mathcal{R}_c^{t_j}|\mathbf{x}_c^{t_j-\Delta}, \theta \right] = \mathrm{e}^{\theta_{\mathcal{M}}^T \mathbf{x}_{c}^{t_j - \Delta}}
\end{equation}
with $\mathbf{x}_c^{t_j - \Delta} = \left[\boldsymbol{\mathcal{D}}_c \text{  } \boldsymbol{\mathcal{M}}_c^{t_j -\Delta}\right]$ combining both temporal and spatial covariates into a single vector.

An Expectation-Maximization strategy, similar to that of \cite{EL11}, is used for fitting the model parameters. This HP-based model is shown to outperform the Susceptible-Exposed-Infected-
Removed (SEIR) model, most usually associated with disease spread forecasting.

\section{Comparisons with other Temporal Point Process approaches}

HPs, and the simpler PPs, have been the most prevalent choice for modeling time event sequences, but some different approaches have been proposed, which occasionally surpassed the performance of HPs in some situations, such as:
\begin{itemize}
    \item Wold Processes: These are the equivalent of a HP in which only the effect of the most recent event is considered in the computation of the intensity function. This Markovian aspect, regardless of the choice of the excitation function, has been shown in \cite{FF18} to surpass the performance of several HP models for estimation of networked processes.
    \item Intensity-Free Learning of Inter-Event Intervals: Another approach which has been recently introduced involves ignoring the intensity function completely, and focusing on modeling the probabilistic distribution of the time intervals among consecutive events. This distribution is modeled after Normalizing Flows \cite{DR15}, which can be summed up as families of distributions with incremental complexities. The approach was introduced in \cite{OS20,OS202}, and was shown to surpass state-of-the-art neural-based HP models in some large-sized datasets.
    \item Continuous-Time Markov Chain: In this model, the marks correspond to states which have fixed rates (intensities) associated to them. The transition time are sampled from these constant intensities. It has been used as a comparison baseline for some HP models, such as \cite{ND15}.
\end{itemize}

\section{Current Challenges for Further Research}
\label{sec: current_challenges}

Regarding the challenges currently tackled by HP researchers, we could mention:
\begin{itemize}
    \item Enriching HP variants (parametric, nonparametric, neural), or blending them with other ML approaches, so as to make them suitable for specific situations. The works with Multi-Armed Bandits \cite{WC20}, randomized kernels \cite{IF20}, graph neural networks for temporal knowledge graphs \cite{ZH20}  and composition of HP-like Point Processes with Warping functions defined over the time event sequences \cite{HX18} can be considered in this category;
    \item Improving the speed of inference or sampling, so as to reduce the time spent in model estimation an aspect which may be critical for some real-world applications. The works of \cite{AH20} in Bayesian mitigation of spatial coarsening, \cite{FZ20} in multi-resolution segmentation for nonstationary Hawkes process using cumulants, \cite{TL19} on thinning of event sequences for accelerating inference steps, \cite{MM203} on the use of Lambert-W functions of improving sequence sampling, \cite{XC20} on perfect sampling are examples of such, and \cite{NP20} on recursive computation of HP moments;
    \item How to properly evaluate and compare HP models among them: While there has been a lot of work regarding proposing new approaches, the comparison among existing models is often biased or incomplete. The works of \cite{HW20} on how to quantify the uncertainty of the obtained models, \cite{SW20} on measuring goodness-of-fit, \cite{MM202} on robust identification of HPs with controlled terms, and \cite{GB20} on the rigorous comparison of networked point process models are among this type of work;
    \item Theoretical guarantees, properties and formulations of specific HP approaches, such as the works of \cite{FC20} on strong mixing, \cite{XG18} on the consistency of some parametric models, \cite{CL20} on elementary derivations of HP momenta, and \cite{KK20, KK202} on field master equation formulation for HPs.
\end{itemize}

\section{Conclusions}

Hawkes Processes are a valuable tool for modeling a myriad of natural and social phenomena. The present work aimed to give a broad view, to a newcomer to the field, of the Inference and Modeling techniques regarding the application of Hawkes Processes in a variety of domains. The parametric, nonparametric, Deep Learning and Reinforcement Learning approaches were broadly covered, as well as the current research challenges on the topic and the real-world limitations of each approach. Illustrative application examples in the modeling of Retweeting behaviour, Earthquake aftershock occurence and COVID-19 spreading were also briefly discussed, for motivating the applicability of Hawkes Processes in both natural and social phenomena.

\section{Acknowledgements}

Any opinions, findings, and conclusions expressed in this work are those of the author and do not necessarily reflect the views of Samsung R\&D Institute Brazil.

\bibliographystyle{siamplain}
\bibliography{references}
\end{document}